\relax
\documentclass[letterpaper]{article} % DO NOT CHANGE THIS
\usepackage{aaai21_mod}  % DO NOT CHANGE THIS
\usepackage{times}  % DO NOT CHANGE THIS
\usepackage{helvet} % DO NOT CHANGE THIS
\usepackage{courier}  % DO NOT CHANGE THIS
\usepackage[hyphens]{url}  % DO NOT CHANGE THIS
\usepackage{graphicx} % DO NOT CHANGE THIS
\usepackage{natbib}  % DO NOT CHANGE THIS AND DO NOT ADD ANY OPTIONS TO IT
\usepackage{caption} % DO NOT CHANGE THIS AND DO NOT ADD ANY OPTIONS TO IT
\usepackage[switch]{lineno}
\newcommand{\RR}{\mathcal{R}} % RR set
\newcommand{\pr}{\mathbb{P}} % probability
\newcommand{\E}{\mathbb{E}} % expectation
\newcommand{\EE}{\mathbb{E}} % expectation
\newcommand{\cE}{\mathcal{E}} % mathcal E
\newcommand{\cD}{\mathcal{D}} % mathcal D
\newcommand{\cO}{\mathcal{O}} % mathcal O
\newcommand{\tcO}{\tilde{\mathcal{O}}} % mathcal O
\newcommand{\cV}{\mathcal{V}} % mathcal V
\newcommand{\cF}{\mathcal{F}} % filtration
\newcommand{\cG}{\cD} % graph

\newcommand{\cH}{\mathcal{H}} 
\newcommand{\cA}{\mathcal{E}} 
\newcommand{\PP}{\mathbb{P}}
\newcommand{\qed}{ $\square$}
\usepackage{amsmath}
\usepackage{dsfont}
\usepackage[noend,ruled,vlined]{algorithm2e}
\usepackage{amssymb}
\newtheorem{theorem}{Theorem}
\newtheorem{problem}{Problem}
\newtheorem{lemma}{Lemma}
\newtheorem{definition}{Definition}

\newcommand{\ind}{\mathds{1}}
\title{Online Learning with Cumulative Oversampling:\\ Application to Budgeted Influence Maximization}
\author{
    Shatian Wang,\textsuperscript{\rm 1}
    Shuoguang Yang,\textsuperscript{\rm 1}
    Zhen Xu,\textsuperscript{\rm 2}
    Van-Anh Truong\textsuperscript{\rm 1}
}
\affiliations{
    \textsuperscript{\rm 1}Department of Industrial Engineering \& Operations Research,\\ Columbia University, New York, NY 10025\\
    \{sw3219,sy2614,vt2196\}@columbia.edu\\
    
    \textsuperscript{\rm 2}Department of Management Science,\\ 
    Fudan University, Shanghai, China\\
    zhen\_xu@fudan.edu.cn\\

}
\begin{document}

\maketitle
% \linenumbers

\begin{abstract}
We propose a cumulative oversampling (CO) method for online learning.
Our key idea is to sample parameter estimations from the updated belief space once in each round (similar to Thompson Sampling), and utilize the cumulative samples up to the current round to construct optimistic parameter estimations that asymptotically concentrate around the true parameters as tighter upper confidence bounds compared to the ones constructed with standard UCB methods.
We apply CO to a novel budgeted variant of the Influence Maximization (IM) semi-bandits with linear generalization of edge weights, whose offline problem is NP-hard. Combining CO with the oracle we design for the offline problem, our online learning algorithm simultaneously tackles budget allocation, parameter learning, and reward maximization. We show that for IM semi-bandits, our CO-based algorithm achieves a scaled regret comparable to that of the UCB-based algorithms in theory, and performs on par with Thompson Sampling in numerical experiments.
\end{abstract}

\section{Introduction}
The \textit{stochastic multi-armed bandit} (MAB) is a classical problem that models the exploration and exploitation trade-off.  There is a slot machine with multiple arms, each following an unknown reward distribution. In each round of a finite-horizon game, an agent pulls one arm and observes its realized reward. The agent aims to maximize the cumulative expected reward; equivalently, to minimize the cumulative regret over all rounds. To do so, she needs to not only learn the reward distributions of all arms by playing each arm a sufficient number of times (explore), but also to use her current estimate of each arm's reward distribution to make good arm selections (exploit). Two widely used methods to address the exploration-exploitation trade-off are Upper Confidence Bound (UCB) \citep{UCB} and Thompson Sampling (TS) \citep{TSempir,TSorig}. UCB-based algorithms maintain estimates on the upper confidence bounds of the mean arm rewards and treat these bounds as proxies for the true mean arm rewards when making decisions. TS-based algorithms maintain a belief over the distributions of the parameters to be learned. In each round, they randomly sample the parameters from the distributions and treat these sampled parameters as proxies for the true parameters when making decisions. After observing feedback, both types of algorithms update empirical beliefs accordingly. 

TS was proposed by \cite{TSorig} more than 80 years ago and has achieved superior empirical performance over other state-of-the-art methods (including UCB) for different variants of MAB \citep{TSempir, TS-Asymp}. However, the theoretical guarantees for TS-based algorithms are limited compared to those of the UCB family, mainly due to the difficulty of controlling deviations from random sampling. In 2012, some progress was made on the theoretical analysis of TS applied to the linear contextual bandit. In this variant, each arm has an associated known $d$-dimensional feature vector and the expected reward of each arm is given by the dot product of the feature vector and an unknown global vector $\theta^* \in \mathbb{R}^d$. \cite{TS-contextual} consider TS as a Bayesian algorithm with a Gaussian prior on $\theta^*$ that is updated and sampled from in each round. They prove a high probability regret bound of $\tilde{O}(d^{3/2}\sqrt{T})$.\footnote{$\tcO$ is a variant of the big $\cO$ notation that ignores all the logarithmic dependencies.}

Following the intuition of \cite{TS-contextual}, \cite{abeille2017linear} show that sampling from an actual Bayesian posterior is not necessary; the same order of regret (frequentist) is achievable as long as TS samples from a distribution that obeys suitable concentration and anti-concentration properties, which can be achieved by \emph{oversampling} the standard least-squares confidence ellipsoid by a factor of $\sqrt{d}$. \cite{NIPS2019_8578} further extend the oversampling approach inspired by \cite{abeille2017linear} to an online dynamic assortment selection problem with contextual information; it assumes a multinomial logit choice model, in which the utility of each item is given by the dot product of a $d$-dimensional context vector and an unknown global vector $\theta^*$. 
Let $K$ denote the number of items to choose for the assortment. Then in each round, their oversampling-based TS algorithm draws a sample set of size $\lceil 1- \frac{\ln K}{\ln(1-1/(4\sqrt{e\pi}))}\rceil \approx 11 \cdot \ln K$ from a least-squares confidence ellipsoid to construct the optimistic utility estimations of the items in the choice set. The optimistic utility estimations are then fed into an efficient oracle which solves for the corresponding optimal assortment. This oversampling idea can be applied to online learning problems whose corresponding offline problems are easy to solve optimally. However, the regret analysis is not extendable to bandits with NP-hard offline problems (detailed in Section \ref{RA}). There is thus a need to design online learning methods that have both superior empirical performance and small theoretical regret for bandits with NP-hard offline problems.

In this paper, we propose such an online learning method that is inspired by the oversampling idea for TS. We apply our new method to a budgeted variant of the Influence Maximization semi-bandits (IM-L) \cite{OIM, contextualIM, IMB, DISBIM, wen2017online}, whose offline problem is NP-hard. 

In IM-L, a social network is given as a directed graph with nodes representing users and edges representing user relationships. For two users Alice and Bob, an edge pointing from Alice to Bob signifies that Bob is a \textit{follower} of Alice. Influence can spread from Alice to Bob (for example, in the form of product adoption). Given a finite horizon consisting of $T$ rounds and a cardinality constraint $K$, an agent selects a \emph{seed set} of $K$ nodes in each round to start an influence diffusion process that typically follows the \emph{Independent Cascade (IC)} diffusion model \cite{Kempe:2003}.  Initially, all nodes in the seed set are activated.  Then in each subsequent time step, each node activated in the previous step has a single chance to independently activate its downstream neighbors with success probabilities equal to the \emph{edge weights}.
Each round terminates once no nodes are activated in a diffusion step. IM-L assumes that the edge weights are initially unknown. The agent chooses seed sets to simultaneously learn the edge weights and maximize the expected cumulative number of activated nodes.  These problems typically assume \emph{edge semi-bandit feedback}; namely, for every node activated during the IC process, the agent observes whether the node's attempts to activate its followers are successful. In this case, we say that the observed \textit{realization} of the corresponding edge is a success; otherwise it is a failure. The agent learns the edge weights using edge semi-bandit feedback. With this feedback structure, IM-L can be cast as combinatorial semi-bandits with probabilistically triggered arms (CMAB-prob) \cite{CUCB}: in each round, a set of arms (as opposed to a single arm) are pulled and the rewards for these pulled arms are observed. Furthermore, pulled arms can probabilistically trigger other arms; the rewards for these other arms are also observed. In IM-L, the arms pulled by the agent in each round are the edges starting from the chosen seed set. The probabilistically triggered arms, arms which are not pulled but their rewards are still observed, are edges starting from nodes that are activated during the diffusion process but not in the seed set. 

When no learning is involved and the edge weights are known, IM-L's corresponding offline problem of finding an optimal seed set of cardinality $K$ is NP-hard \cite{Kempe:2003}. Since the expected number of activated nodes as a function of seed sets is monotone and submodular, the greedy algorithm achieves an approximation guarantee of $1-1/e$ if the function values can be computed exactly \cite{Nemhauser1978}. However, because computing this function is \#P-hard, it requires simulations to be estimated \citep{Chen:2010:SIM:1835804.1835934}. 

Existing learning algorithms for IM-L thus all assume the existence of an $(\alpha,\beta)$-approximation oracle that returns a seed set whose expected reward is at least $\alpha$ times the optimal with probability at least $\beta$, with respect to the input edge weights and cardinality constraint. These learning algorithms use UCB- or TS-based approaches in each round to estimate the edge weights and subsequently feed these updated estimates to the oracle, producing a seed set selection \citep{OIM, contextualIM, IMB, DISBIM, wen2017online}. \cite{wen2017online} is the first to scale up the learning process by assuming linear generalization of edge weights. That is, each edge has an associated $d$-dimensional feature vector that is known by the agent, and the weight on each edge is given by the dot product of the feature vector and an unknown global vector $\theta^* \in \mathbb{R}^d$. Let $n$ denote the number of nodes and $m$ denote the number of edges in the input directed graph. With this assumption, \cite{wen2017online} propose a UCB-based learning algorithm for IM-L that achieves a scaled regret of $\tcO(dnm\sqrt{T})$. This improves upon the existing regret bound in \cite{CUCB} that is linearly dependent on $1/p^*$, where $p^*$ is the minimum observation probability of an edge. $1/p^*$ can be exponential in $m$, the number of edges.

% Although IM-L has been extensively studied, there are still gaps to be filled. Notably, there are no online learning algorithms for IM-L with both superior empirical performance and small theoretical regret. 
For IM-L, TS-based algorithms often significantly outperform UCB-based ones in simulated experiments \cite{TSempir,Hyk2019ThompsonSF, TS-Asymp}, but few regret analysis exists for TS-based algorithms.\footnote{\cite{Hyk2019ThompsonSF} derive a regret bound for a TS-based algorithm applied to CMAB-prob. Their regret bound still depends linearly on $1/p^*$, which can be exponential in $m$.}
% Even with linear generalization, extending the UCB analysis of \cite{wen2017online} to TS-based algorithms is highly non-trivial; the reason will be detailed later in this paper.

\noindent \textbf{Our contribution} We propose a novel cumulative oversampling method (CO) that can be applied to IM-L and potentially to many other bandits with NP-hard offline problems. CO is inspired by the oversampling idea for Thompson Sampling in \cite{abeille2017linear} and \cite{NIPS2019_8578}, but requires significantly fewer samples compared to \cite{NIPS2019_8578}. Exactly one sample needs to be drawn from a least-squares confidence ellipsoid in each round. Our key idea is to utilize all the samples up to the current round to construct optimistic parameter estimations. 
In practice,
CO is similar to TS with oversampling in the initial learning rounds.
As the number of rounds increases, the optimistic parameter estimations serve as tighter upper confidence bounds compared to the ones constructed with UCB-based methods.

We apply CO to a budgeted variant of IM-L which we call \emph{Budgeted Influence Maximization Semi-Bandits with linear generalization of edge weights (Lin-IMB-L)}. 
In it, each node charges a different commission to be included in a seed set. Unlike IM-L that imposes a fixed cardinality constraint $K$ for each round, Lin-IMB-L assumes that there is a global budget $B$ that needs to be satisfied in expectation over a finite horizon of $T$ rounds. The agent needs to allocate the budget to $T$ rounds as well as learning edge weights and maximizing cumulative reward. For this problem, we analyze its corresponding offline version and propose the first $(\alpha,\beta)$-approximation oracle for it. To develop this oracle, we extend the state-of-the-art \emph{Reverse Reachable Sets (RRS)} simulation techniques for IM \citep{DBLP:journals/corr/abs-1212-0884, Tang:2014:IMN:2588555.2593670, Tang:2015:IMN:2723372.2723734} to accurately estimate the reward of seed sets of any size. 
We combine our cumulative oversampling method with our oracle into an online learning algorithm for Lin-IMB-L. We prove that the scaled regret of our algorithm is in the order of $\tcO(dnm\sqrt{T})$, which matches the regret bound for the UCB-based algorithm for IM-L with linear generalization of edge weights proved by \cite{wen2017online}. We further conduct numerical experiments on two Twitter subnetworks and show that our algorithm performs on par with Thompson Sampling and outperforms all UCB-based algorithms by a large margin with or without perfect linear generalization of edge weights.

\section{Budgeted IM Semi-Bandits} \label{PF}
We mathematically formulate our new budgeted IM semi-bandits problem in this section.
We model the topology of a social network using a directed graph $\mathcal{D = (V, E)}$. 
Each node $v \in \mathcal{V}$ represents a user, and an arc (directed edge) $(u,v) \in \mathcal{E}$ indicates that user $v$ is a \textit{follower} of user $u$ in the network and influence can spread from user $u$ to user $v$.
For each arc $e = (u,v)$, we use $\bar{w}(e) \in [0,1]$ to denote the \textit{edge weight on $e$}. 
There are in total $n$ nodes and $m$ arcs in $\mathcal{D = (V, E)}$.
Throughout the text, we refer to the function $\bar{w}: \cA \mapsto [0,1]$ as \textit{true edge weights}. 

Once a seed set $S \subseteq \mathcal{V}$ is selected, influence spreads in the network from $S$ following the \textit{Independent Cascade Model (IC)} \citep{Kempe:2003}. IC specifies an influence spread process in \textit{discrete time steps}. In the initial step, all seeded users in $S$ are \textit{activated}. In each subsequent step $s$, each user activated in step $s-1$ has a single chance to activate its followers, or \textit{downstream neighbors}, with success rates equal to the corresponding edge weights. This process terminates when no more users can be activated. We can equivalently think of the IC model as flipping a biased coin on each edge and observing connected components in the graph with edges corresponding to positive flips \citep{Kempe:2003}. More specifically, after the influencers in the seed set $S$ are activated,  the environment decides on the binary weight function $\mathbf{w}$ by independently sampling $\mathbf{w}(e) \sim \text{Bern}(\bar{w}(e))$ for each $e\in \cE$. 
A node $v_2 \in \mathcal{V}\backslash S$ is \textit{activated} by a node $v_1 \in S$ if there exists a directed path $e_1, e_2, ..., e_l$ from $v_1$ to $v_2$ such that $\mathbf{w}(e_i) = 1$ for all $i = 1, ..., l$. Let $I(S, \mathbf{w}) = \{v\in V| v\in S \text{ or } v \text{ is activated by a node }u\in S \text{ under } \mathbf{w}\}$ be the set of nodes activated during the IC process given seed set $S$. We denote the expected number of activated nodes given seed set $S$ and edge weights $\bar{w}$ by $f(S, \bar{w})$, i.e., $f(S, \bar{w}):= \E\left(|I(S, \mathbf{w})|\right)$, and refer to the realization of $\mathbf{w}(e)$ as the \textit{realization of edge $e$}.

Below, we formally define our \emph{Budgeted Influence Maximization Semi-Bandits with linear generalization of edge weights (Lin-IMB-L)}. In it, an agent runs an influencer marketing campaign over $T$ rounds to promote a product in a given social network $\cD$. The agent is aware of the structure of $\cD$ but initially does not know the edge weights $\bar{w}$. In each round $t$, it activates a seed set $S_t$ of influencers in the network by paying each influencer $u \in S_t$ a fixed commission $\mathbf{c}(u) \in \mathbb{R}^+$ to promote the product. 
Influence of the product spreads from $S_t$ to other users in the network in round $t$ according to the IC model. For each round $t$, we assume that the influence spread process in this round terminates before the next round $t+1$ is initiated.
The total cost of selecting seed set $S_t$ is denoted by $\mathbf{c}(S_t) = \sum_{u \in S_t}\mathbf{c}(u)$. A exogenous budget $B$ is given at the very beginning. The campaign selects seed sets with the constraint that \textit{in expectation}, the cumulative cost over $T$ rounds cannot exceed $B$, where the expectation is over possible randomness of $S_t$, since it can be returned by a randomized algorithm. The goal of the agent is to maximize the expected total reward over $T$ rounds.

As in \cite{wen2017online}, we assume a \textit{linear generalization} of $\bar{w}$. That is, for each arc $e \in \cE$, we are given a \textit{feature vector} $\mathbf{x}_e \in \mathbb{R}^d$ that characterizes the arc. Also, there exists a vector $\theta^* \in \mathbb{R}^{d}$ such that the edge weight on arc $e$, $\bar{w}(e)$, is closely approximated by $ \mathbf{x}_e^\top \theta^*$. $\theta^*$ is initially unknown. The agent needs to learn it over the finite horizon of $T$ rounds through  \textit{edge semi-bandit feedback} \citep{CUCB, wen2017online}. That is, for each edge $e = (u,v) \in \cE$, the agent observes the realization of $\mathbf{w}(e)$ in round $t$ if and only if $u \in I(S_t, \mathbf{w})$, i.e., the head of the edge was activated during the IC process in round $t$. We refer to the set of edges whose realizations are observed in round $t$ as the set of \textit{observed edges}, and denote it as $\cE_t^o$. Depending on whether or not the tail node of an observed edge is activated, the realization of the edge can be either a success ($\mathbf{w}(e) = 1$), or a failure ($\mathbf{w}(e) = 0$). 

\begin{problem}
	\textbf{Budgeted Influence Maximization Semi-Bandits with linear generalization of edge weights (Lin-IMB-L)}\\ 
	Given a social network $\mathcal{D = (V, E)}$, edge feature vectors $\mathbf{x}_e \in \mathbb{R}^d \; \forall e\in \cE$, cost function $\mathbf{c}: \mathcal{V} \rightarrow \mathbb{R}^+$, budget $B \in \mathbb{R}^+$, finite horizon $T \in \mathbb{Z}^+$; assume $\bar{w}(e) = \mathbf{x}_e ^\top \theta^*$ for some unknown $\theta^* \in \mathbb{R}^d$, and that the agent observes edge semi-bandit feedback in each round $t = 1, ..., T$. In each round $t$, adaptively choose $S_t \subseteq \mathcal{V}$ so that \begin{equation}\label{eq:bdgt-cnstr}
	\begin{split}
	\{S_t\}_{t=1}^T \in \arg \max \Big \{&\E\Big[ \sum_{t=1}^T f(S_t, \bar{w})\Big]:\\ 
	&\E\Big[ \sum_{t=1}^T \mathbf{c}(S_t)\Big] \leq B\Big \}.
	\end{split}
	\end{equation}
	\end{problem}
	
	Lin-IMB-L presents three challenges. First, the agent needs to learn the edge weights through learning $\theta^*$ over a finite time horizon. Second, the agent needs to allocate the budget to individual rounds. Third, the agent needs to make a good seeding decision in each round that balances exploration (gather more information on $\theta^*$) and exploitation (maximize cumulative reward using gathered information). Our online learning algorithm uses cumulative oversampling (CO) to construct an optimistic (thus exploratory) estimate $\tilde u_t: \cA \mapsto [0,1]$ on the edge weights $\bar{w}$ in each round $t$ using the edge semi-bandit feedback gathered so far (thus exploitative). It then feeds $\tilde u_t$ together with the budget $b$ allocated to the current round to an approximation oracle in order to decide on a seed set for the current round. In the next section, we propose the first such approximation oracle and prove its approximation guarantee. Then in Section \ref{OL}, we detail our online learning algorithm and the CO method behind it.

	\section{Approximation Oracle} \label{offlineORACLEs} 
	Assume that we have an estimate $w$ on the edge weights $\bar{w}$ and an expected budget $b$ for the current round, an important subproblem of Lin-IMB-L is that in each round, we want to choose a seed set that maximizes the expected reward with respect to $w$ while respecting the budget constraint $b$. We refer to this subproblem as IMB and formally define it below. 
		\begin{problem} \label{problem:IMB}\textbf{IMB}\\ Given network $\cD=(\mathcal{V}, \cE)$, budget $b$, cost function $\mathbf{c}$, edge weights $w$, find $S \subseteq \mathcal{V}$ such that $\E(\mathbf{c}(S)) \leq b$, and $\E(f(S, w))$ is maximized. The expectations are over possible randomness of $S$, since $S$ can be returned by a randomized algorithm.
	    \end{problem}

IMB is NP-hard (see Appendix \ref{apd:nphardness-proof} for a reduction from the set cover problem to it).
Below, we propose an approximation oracle for IMB. We refer to it as {\tt{ORACLE-IMB}}. Further note that $f(\cdot, w)$ is \#-P hard to compute \cite{Chen:2010:SIM:1835804.1835934}, and we thus need efficient simulation-based methods to accurately estimate it with high probability. We defer the estimation of $f(\cdot, w)$ to Appendix \ref{apd:simulation}, where we detail how to modify our oracle to incorporate the estimation of $f(\cdot, w)$ and prove that the resulting algorithm's $(\alpha, \beta)$-approximation ratio. We refer to the modified oracle as {\tt{ORACLE-IMB-M}}.

\begin{algorithm}
\SetAlgoLined
  \KwData{$\cD = (\cV, \cE), b, \mathbf{c}, w$}
  \KwResult{$S\subseteq \mathcal{V}$}
 initialization: $S_0 = \emptyset$\;
 \For{$i = 1, 2, ..., n$}{Compute $v_i = {\arg\max}_{v \in \mathcal{V}\backslash S_{i-1}} \frac{f(S_{i-1}\cup\{v\}, w)-f(S_{i-1}, w)}{\mathbf{c}(v)}$\;
 Set $S_i = S_{i-1} \cup \{v_i\}$\;
 \If{$\mathbf{c}(S_i)>B$}{
 Set $S_-=S_{i-1}, S_+=S_i$\;
 Break}}
 Solve the following LP to get an optimal solution $(p^*, q^*)$:\\
$\;\;\max \; p \cdot f(S_-, w) + q \cdot f(S_+, w)$\; $\;\;\text{s.t}\; p\cdot\mathbf{c}(S_-) + q\cdot \mathbf{c}(S_+) \leq b; \; p+q =1; \; p,q \geq 0$\; 
Sample $S$ from $\{S_-, S_+\}$ with probability distribution $(p^*,q^*)$
 \caption{{\tt{ORACLE-IMB}}}
\end{algorithm}

We have the following approximation guarantee for {\tt{ORACLE-IMB}} (proved in Appendix \ref{apd:offline_thm2-proof}). Note that the existing approximation algorithm for budgeted monotone submodular function maximization with a deterministic budget needs to evaluate all seed sets of size up to $3$ to achieve an $1-1/e$-approximation \cite{budgetedSubm}. Our {\tt{ORACLE-IMB}} does not have this computationally expensive partial enumeration step. With an expected budget, we have the same approximation guarantee.
	
	\begin{theorem} \label{offline_thm2}
		For any IMB instance, $\E(f(S^{ora}, w)) \geq (1-1/e)\E(f(S^*,  w))$, where $S^{ora}$ is the seed set returned by {\tt{ORACLE-IMB}} and $S^*$ is the seed set selected by an optimal algorithm. %Thus, $E\left[ \sum_{t=1}^T f(S^{B}_w(t), w)\right]  \geq (1-1/e)E\left[ \sum_{t=1}^T f(S^*_w(t),  w)\right] = (1-1/e) E({\tt{OPT}(T)})$.
	\end{theorem}
	
\section{Online Learning Algorithm for Lin-IMB-L} \label{OL}

To utilize Thompson Sampling to solve Lin-IMB-L, one could maintain a belief on the distribution of $\theta^*$ and sample a $\tilde \theta$ from the updated belief in each decision round, treating it as the nominal mean when making the seeding decision.  Thompson Sampling, while demonstrating superior performance in experiments, is hard to analyze, mainly due to the difficulty in controlling the deviations resulting from random sampling.

\cite{abeille2017linear} show that for linear contextual bandits, sampling from an actual Bayesian posterior is not necessary, and the same order of regret (frequentist) is achievable as long as the the distribution TS samples from follows suitable concentration and anti-concentration properties, which can be achieved by \emph{oversampling} the standard least-squares confidence ellipsoid by a factor of $\sqrt{d}$. The oversampling step is used to guarantee that the estimates have a constant probability of being optimistic. \cite{NIPS2019_8578} extend this idea to a dynamic assortment optimization problem with MNL choice models. Their oversampling-inspired TS algorithm uses $11 \cdot \ln(K)$ samples from the least-squares confidence ellipsoid in each round to construct the optimistic utility estimates of the items in the choice set, where $K$ is the number of items in the assortment. For both linear contextual bandits and the dynamic assortment optimization with MNL choice models, the optimal ``arm'' with respect to the parameter estimates can be efficiently computed. 

However, oversampling a constant number of samples in each round is insufficient to guarantee a small regret for bandits with NP-hard offline problems, for which there exist only $(\alpha,\beta)$-approximation oracles returning an $\alpha$-approximation arm with probability at least $\beta$. We postpone the explanations of the challenges to Section \ref{RA}. 

We propose an alternative \textit{cumulative oversampling (CO)} method that can be applied to Lin-IMB-L and potentially to other bandits with NP-hard offline problems to obtain bounded small regrets and superior empirical performance. 

Under CO, in each round $t$, we sample exactly one $\tilde{\theta}_t$ from the multivariate Gaussian distribution ${\mathcal{N}} ( \theta_{t}, v^2\alpha_t^2\mathbf{M}_{t-1}^{-1})$ where $\theta_{t}$ is the regularized least squares estimator of $\theta^*$, $v \in \mathbb{R}^+$ is a hyper-parameter, $\mathbf{M}_{t-1}$ is the corresponding design matrix, and 
\begin{equation} \label{eq:alpha_t}\alpha_t := \sqrt{d \ln (1+  \frac{tm}{ d} ) + 4\ln t} + D,\end{equation}
with $D$ being a known upper bound for $\|\theta^*\|_2$.

For any real vector $y \in \mathbb{R}^d$ and positive semi-definite matrix $\mathbf{M} \in \mathbb{R}^{d\times d}$ let $||y||_\mathbf{M} = \sqrt{y^\top \mathbf{M} y}$ be a norm of $y$ weighted by $\mathbf{M}$.

Define 
\begin{equation} \label{eq:wtildeconcentration}
\sigma_\tau(e) = \frac{\tilde w_\tau(e) - \mathbf{x}_e^\top \theta_\tau}{ \alpha_\tau \|\mathbf{x}_e \|_{\mathbf{M}_{\tau-1}^{-1}} }.
\end{equation}
We construct the edge weights estimate $\tilde w_t$ for the current round recursively as follows. For each $e\in \cE$, let
$$\tilde w_t(e) = \max \Big(\mathbf{x}_e^\top\tilde \theta_t, \; \mathbf{x}_e^\top\theta_t + \sigma_{t-1}(e)\alpha_t \|\mathbf{x}_e \|_{\mathbf{M}_{t-1}^{-1}}\Big).$$

And we define $\tilde u_t(e)$ as the projection of $\tilde w_t(e)$ onto $[0,1]$. We then feed $\tilde u_t$ into our seeding oracle {\tt{ORACLE-IMB}}. The details of the resulting algorithm is summarized below.
\begin{algorithm} 
\SetAlgoLined
  \KwData{digraph $\cD=\mathcal{(V, E)}$, node costs $\mathbf{c}: \mathcal{V} \mapsto \mathbb{R}^+$, edge feature vectors $\mathbf{x}_e \in \mathbb{R}^d\; \forall e\in \cE$, number of rounds $T$, known upper bound $D$ for $\|\theta^*\|$, hyper-parameter $v \in \mathbb{R}^+$.}
  \KwResult{$S_t\subseteq \mathcal{V}, t = 1, ..., T$.}
 Initialization: $\mathbf{M}_{-1} = \mathbf{M}_{0} = I \in \mathbb{R}^{d\times d}$, $\mathbf{B}_{0} = 0 \in \mathbb{R}^d$, $\tilde w_0(e) = -\infty \; \forall e \in \cE, \alpha_0 = 1$\;
 \For{{$t = 1, 2, ..., T$}}{Set $\theta_{t} = \mathbf{M}_{t-1}^{-1}\mathbf{B}_{t-1}$\;
 Compute $\alpha_t = \sqrt{d \ln (1+  \frac{tm}{ d} ) + 4\ln t} + D$\;
 Sample $\tilde{\theta}_{t}$ from ${\mathcal{N}} ( \theta_{t}, v^2\alpha_t^2\mathbf{M}_{t-1}^{-1})$\;
 \For{{$e\in \cE$}}{
 Compute $\sigma_{t-1}(e) = \frac{\tilde w_{t-1}(e) - \mathbf{x}_e^\top \theta_{t-1}}{ \alpha_{t-1} \|\mathbf{x}_e \|_{\mathbf{M}_{t-2}^{-1}} }$,\\
 ${\displaystyle \tilde{w}_t(e)= \mathbf{x}_e^\top\tilde \theta_t \vee \Big(\mathbf{x}_e^\top\theta_t + \sigma_{t-1}(e)\alpha_t \|\mathbf{x}_e \|_{\mathbf{M}_{t-1}^{-1}}\Big)}$,\\ 
 $\tilde u_t(e) = \text{Proj}_{[0,1]} \tilde w_t(e) $\;
$S_t \leftarrow \text{\tt{ORACLE-IMB}}(\cD, B/T, \mathbf{c}, \tilde{u}_t)$\;}
 Select seed set $S_t$ and observe semi-bandit edge activation realizations\;
 Update $\mathbf{M}_{t} = \mathbf{M}_{t-1} + \sum_{e\in \mathcal{E}_t^o}\mathbf{x}_e\mathbf{x}_e^\top$, $\mathbf{B}_{t} = \mathbf{B}_{t-1} + \sum_{e\in \mathcal{E}_t^o}\mathbf{x}_e y_e^t$, \\ $ \; $where $\mathcal{E}_t^o$ is the set of edges whose realizations are observed, and $y_e^t \in \{0,1\}$ is the realization of edge $e$ in round $t$.
 }
 \caption{Cumulative Oversampling (CO)}
 \label{alg:TS_IMcB}
\end{algorithm}

CO practically preserves the advantages of both TS- and UCB-based algorithms:
CO is similar to TS with oversampling in the initial learning rounds, whose superior empirical performance over other state-of-art methods such as UCB has been shown. 
As the number of rounds increases, the weight estimate $\tilde u_t$ serves as a tighter upper confidence bound that achieves smaller regrets. This CO method sheds light on designing algorithms with small regret guarantees and superior empirical performance for other NP-hard problems. The exact proof for the asymptotic concentration of the estimators constructed using the cumulative samples might differ from problem to problem, but the general regret analysis outline shall be fairly similar to the one presented in the next section.

\section{Regret Analysis} \label{RA}
We first explain why the existing oversampling method does not alleviate the challenges in regret analysis of TS-based algorithms for bandits with NP-hard offline problems in general. We then present the regret result of applying CO to Lin-IMB-L and provide a proof sketch.

For any bandits whose offline problem can be solved to  optimality efficiently, we can analyze the regret as follows. Use $S^*(w)$ to denote the optimal action given parameter $w$. In each round $t$, action $S^*(\tilde w_t)$ is taken by the online learning algorithm given the parameter estimate $\tilde w_t$. Let $\bar{w}$ be the true parameter. Use $f(S, w)$ to denote the expected reward of action $S$ under parameter $w$. The expected reward for round $t$ can thus be represented as $\E(f(S^*(\tilde w_t), \bar{w}))$. The optimal expected reward for each round is $E(f(S^*(\bar{w}), \bar{w}))$. The preceding expectations are over the possible randomness of $S^*(w)$, since the optimal action can be randomized.

The expected regret for round $t$ is defined as 
$$\E(R_t) = \E(f(S^*(\bar{w}), \bar{w}) - f(S^*(\tilde w_t), \bar{w})).$$ The cumulative regret $R(T)$ over $T$ rounds is defined as the sum of $\E(R_t)$'s.

The expected round regret $\E(R_t)$ is usually decomposed as follows:
\begin{equation}\label{bound:diff}
\begin{split}
    \E(R_t)
    &= \underbrace{\E(f(S^*(\bar w),\bar w)  - f(S^*(\tilde w_t),\tilde w_t))}_{R_t^1}\\ 
    & + \underbrace{\E(f(S^*(\tilde w_t),\tilde w_t) - f(S^*(\tilde w_t),\bar w))}_{R_t^2}.
\end{split}
\end{equation} 
While bounding $R_t^2$ is relatively straightforward using standard bandit techniques, bounding $R_t^1$ requires more careful analysis. Intuitively, however, when $\tilde w_t$ and $\bar w$ are close enough, the difference between their corresponding optimal rewards is likely small as well. Indeed, this has been shown for the stochastic linear bandits and the assortment optimization settings using the constant optimistic probability achieved with traditional oversampling \cite{abeille2017linear,NIPS2019_8578}.

On the other hand, when the underlying problem is NP-hard, an $(\alpha,\beta)$-approximation oracle has to be used in the learning algorithm. It takes the parameter estimate $\tilde w_t$ as input and returns an action $S_t$ such that $f(S_t, \tilde w_t) \geq \alpha \cdot f(S^*(\tilde w_t), \tilde w_t)$ with probability at least $\beta$. As a result, a scaled regret analysis is performed instead: one is now interested in bounding the $T$-round $\eta$-scaled regret 
\begin{equation}\label{eq:scaledregret_decomp}
R^\eta(T) = \sum_{t = 1}^T \E(R^\eta_t),\end{equation} where $\eta = \alpha \beta$ and
\begin{equation} \label{bound:eta_diff}
\begin{split}
     \E(R^\eta_t) &= \E \left[f(S^*(\bar w),\bar w)  -  \frac{f(S_t,\bar w)}{\eta} \right],\\
     & = \underbrace{\E \left[f(S^*(\bar w),\bar w)  -  \frac{f(S_t,\tilde w_t)}{\eta} \right]}_{R_t^1}\\ &+ \frac{1}{\eta}\underbrace{\E \left[f(S_t,\tilde w_t) - f(S_t,\bar w)\right]}_{R_t^2}.
\end{split}
 \end{equation}
 Again, the difficulty mainly arises in bounding $R_t^1$. Use $S^\eta(w)$ to denote any solution such that $\E[f(S^\eta(w), w)] \geq \eta \cdot \E[f(S^*(w), w)]$. By definition of $S^\eta(w)$ and the property of the $(\alpha,\beta)$-approximation oracle, we can establish the following two upper bounds for $R_t^1$:
 \begin{align}
     R_t^1  \leq & \,\,\,\, \E\left [f(S^\eta(\bar w),\bar w) - f(S_t,\tilde w_t) \right ] /\eta \label{bound:scaled_diff_1}, \\
     R_t^1
     \leq & \,\,\,\, \E[ f(S^*(\bar w),\bar w) - f(S^*(\bar{w}),\tilde w_t)]. \label{bound:scaled_diff_2}
 \end{align}
 
In Eq.\eqref{bound:scaled_diff_1}, even when $\tilde w_t = \bar w$, $R_t^1$ does not necessarily diminish to $0$. This is because $\E[f(S^\eta(\bar w), \bar w)] \geq \eta \cdot \E[f(S^*(\bar w), \bar w)]$ and $\E[f(S_t, \tilde w_t)] \geq \eta \cdot \E[f(S^*(\tilde w_t), \tilde w_t)]$ do not guarantee $\E[f(S^\eta(\bar w), \bar w)] = \E[f(S_t, \tilde w_t)]$. To bound the RHS of Eq. \eqref{bound:scaled_diff_2} is also challenging, because all the observations gathered by the agent is under action $S_t$ and $\bar w$. Losing the dependency on $S_t$ means the observations under $S_t$ cannot be utilized to construct a good upper bound. 

On the other hand, with CO, we can prove that as $t$ increases, the probability that $\tilde w_t$ is an (point-wise) upper bound for $\bar w$ increases fast enough. As a result, if $f(S,\cdot)$ is monotone increasing, then the RHS of Eq. \eqref{bound:scaled_diff_2} can be upper bounded by 0 with a higher probability as $t$ increases. In essence, CO is similar to TS with oversampling in the initial rounds, but asymptotically its analysis is more similar to the analysis for UCB-based algorithms.

Below, we present the regret analysis of CO for IM with linear generalization of edge weights. Prior to this work, only regret bounds for UCB algorithms have been established \cite{wen2017online} to the best of our knowledge.
\begin{theorem}\label{thmTS-CO}
Let $\eta = 1-1/e$, 
the $T$-round $\eta$-scaled regret of Algorithm \ref{alg:TS_IMcB} is
\begin{equation*}
\begin{split}
    R^{\eta}(T) &\leq \frac{(\alpha_T + \beta_T)nm}{\eta} \sqrt{  \frac{dT\ln (1+ \frac{mT}{d})}{\ln 2}}\\
    & + n \Big ( \frac{4m\sqrt{\pi }  e^{1/2v^2}}{v}  + \frac{\pi^2}{3}\Big ).
\end{split}
\end{equation*}
where $\alpha_t$ is defined in \eqref{eq:alpha_t} and $$\beta_t = v\alpha_t(\sqrt{2\ln 2t} +\sqrt{2\ln m + 4\ln t}).$$
\end{theorem}
\textit{Proof sketch (see Appendix \ref{apd:online} for full proof):} for each round $t$, we define two favorable events $\xi_t$, $\delta_t$ (and their complements $\bar{\xi}_t$, $\bar{\delta}_t$):
$$\xi_t := \{|x_e^\top\theta_t - \bar{w}(e)| \leq \alpha_t \|\mathbf{x}_e\|_{\mathbf{M}_{t-1}^{-1}}\; \forall e\in \cE\},$$
$$\delta_t := \{|\tilde w_t(e) - x_e^\top\theta_t| \leq \beta_t \|\mathbf{x}_e\|_{\mathbf{M}_{t-1}^{-1}} \; \forall e\in \cE\}.$$

We can represent the cumulative scaled regret $R^\eta(T)$ as the sum of expected round scaled regrets as in \eqref{eq:scaledregret_decomp} and \eqref{bound:eta_diff} with $S^* = S^*(\bar{w})$ being the seed set selected by an optimal (randomized) oracle for IMB with input edge weights $\bar{w}$ and budget $B/T$ (for detailed explanation, see Appendix \ref{apd:regretdecomp}).

Decomposing $\E(R_t^\eta)$ by conditioning on $\xi_t$ and $\bar{\xi}_t$, and using the naive bound $R_t^\eta \leq n$, we get
\begin{equation*}
\begin{split}
\EE[R_t^\eta] &\leq \underbrace{\EE [ f(S^*,\bar w) - \frac{1}{\eta} f(S_t,\tilde u_t) | \xi_t ] }_{Q_1} \cdot \PP(\xi_t )\\ 
& + \frac{1}{\eta} \underbrace{\EE[ f(S_t,\tilde u_t) -  f(S_t, \bar w) | \xi_t] }_{Q_2} \cdot \PP(\xi_t ) + n\cdot \PP(\bar \xi_t).
\end{split}
\end{equation*}
By standard linear stochastic bandits techniques in \cite{Abbasi-Yadkori:2011}, we can obtain $$\PP(\bar{\xi}_t) \leq 1/t^2.$$

For any two edge weights functions $u: \cE \mapsto [0,1]$ and $w: \cE \mapsto [0,1]$, if $u(e) \geq w(e) \; \forall e\in \cE$, then we write $u \geq w$.
We show $$Q_1 \leq n \big (1-\PP(\tilde u_t \geq \bar w \;| \xi_t) \big )= n \big (1-\PP(\tilde w_t \geq \bar w \;| \xi_t) \big ).$$
Due to the way we construct $\tilde w_t$, we can lower bound $\PP(\tilde w_t \geq \bar w \;| \xi_t)$ by $$1- m \PP \Big ( \max_{j=1,\cdots,t}Z_j \leq  \frac{1}{v}\Big ) \geq 1- m \Big ( 1-\frac{v}{4\sqrt{\pi } } e^{-1/2v^2} \Big )^t.$$

As for $Q_2$, we can further decompose it by conditioning on $\delta_t$ and $\bar{\delta}_t$. We prove that $$\PP(\bar{\delta}_t|\xi_t)\leq 1/t^2.$$ This result together with a lemma adapted from \cite{wen2017online} allow us to upper bound $Q_2$ by
$$\EE \Big [\sum_{v\in \cV \setminus S_t}\sum_{e\in\cE_{S_t,v}} \mathds{1}\{O_t(e)\} | \tilde u_t(e) - \bar w(e) | \Big  |\delta_t, \xi_t  \Big] \PP(\delta_t|\xi_t) + \frac{n}{t^2},$$ where $\cE_{S_t,v}$ is the set of edges that are relevant to whether of not $v$ would be activated given seed set $S_t$ (defined in Appendix \ref{apd:mor}), $O_t(e)$ is the event that edge $e$'s realization is observed given seed set $S_t$ and edge weights $\bar{w}$ in round $t$.

By definition of $\delta_t, \xi_t, \tilde u_t$, we can upper bound $|\tilde u_t(e) - \bar w(e)|$ by $(\alpha_t+\beta_t)\|\mathbf{x}_e\|_{\mathbf{M}_{t-1}^{-1}}$ under $\delta_t, \xi_t$. 

Combining the results above, we have
\begin{equation*}
\begin{split}
\E(R^\eta_t) & \leq nm\Big ( 1-\frac{v}{4\sqrt{\pi } } e^{-1/2v^2} \Big )^t+\frac{2n}{t^2}\\
& \;\;\; + \frac{\alpha_t+\beta_t}{\eta}\cdot  \PP(\delta_t, \xi_t) \cdot\\
& \;\;\; \EE \Big[\sum_{v\in \cV \setminus S_t}\sum_{e\in\cE_{S_t,v}} \mathds{1}\{O_t(e)\}\sqrt{\mathbf{x}_e^\top \mathbf{M}_{t-1}^{-1} \mathbf{x}_e} \Big | \delta_t, \xi_t \Big]
\end{split}
\end{equation*}

Since the inside of the conditional expectation in the last line above is always non-negative, we can remove the conditioning and upper bound $\E(R_t^\eta)$ simply by 
\begin{equation*}
\begin{split}
&nm\Big ( 1-\frac{v}{4\sqrt{\pi } } e^{-1/2v^2} \Big )^t+\frac{2n}{t^2}\\
& \;\;\; + \frac{\alpha_t+\beta_t}{\eta}\EE \Big[\sum_{v\in \cV \setminus S_t}\sum_{e\in\cE_{S_t,v}} \mathds{1}\{O_t(e)\}\sqrt{\mathbf{x}_e^\top \mathbf{M}_{t-1}^{-1} \mathbf{x}_e}\Big].
\end{split}
\end{equation*}

The sum of the first line above over all rounds can be upper bounded by 
$$n \Big ( \frac{4m\sqrt{\pi }  e^{1/2v^2}}{v}  + \frac{\pi^2}{3}\Big )$$
using known results about geometric series and the Basel problem. 

The sum of the second line above over all rounds can be upper bounded by $$\frac{(\alpha_T+\beta_T)nm}{\eta}\sqrt{\frac{d T\ln (1+ \frac{Tm}{d})}{\ln 2}},$$ following standard linear stochastic bandits techniques detailed in \cite{wen2017online}.  \qed

\section{Numerical Experiments} \label{NE}
We conduct numerical experiments on two Twitter subnetworks. The first subnetwork has 25 nodes and 319 directed edges, and the second has 50 nodes and 249 directed edges. We obtain the network structures from \cite{SNAP}, and construct node feature vectors using the {\tt{node2vec}} algorithm proposed in \cite{DBLP:journals/corr/GroverL16}. We then use the element-wise product of two node features to get each edge feature vector. We adopt this setup from \cite{wen2017online}. For the 25-node network, we hand-pick a $\theta^*$ vector so that the edge weight obtained by taking the dot product between each edge feature vector and this $\theta^*$ falls between $0.01$ and $0.15$. Thus we have a perfect linear generalization of edge weights. For the 50-node experiment, we randomly sample an edge weight from {\tt{Unif}}(0,0.1) for each edge. As a result, it is unlikely that there exists a vector $\theta^*$ that perfectly generalizes the edge weights. 

For each subnetwork, we compare the performance of CO with three other learning algorithms, 1) TS assuming linear generalization 2) UCB assuming linear generalization and 3) CUCB assuming \textit{no} linear generalization \cite{CUCB}. We set $T = 5,000$, $d = 10$, and $B = 10,000$ and use {\tt{ORACLE-IMB}-M} as the seeding oracle. We perform 500 rounds of random seeding and belief updates to warm start the campaign. The information gathered during this pre-training phase can be thought of as the existing social network data the campaigner can use to form a prior belief on the parameters.

\begin{figure}
    \centering
    \includegraphics[width=0.8\linewidth]{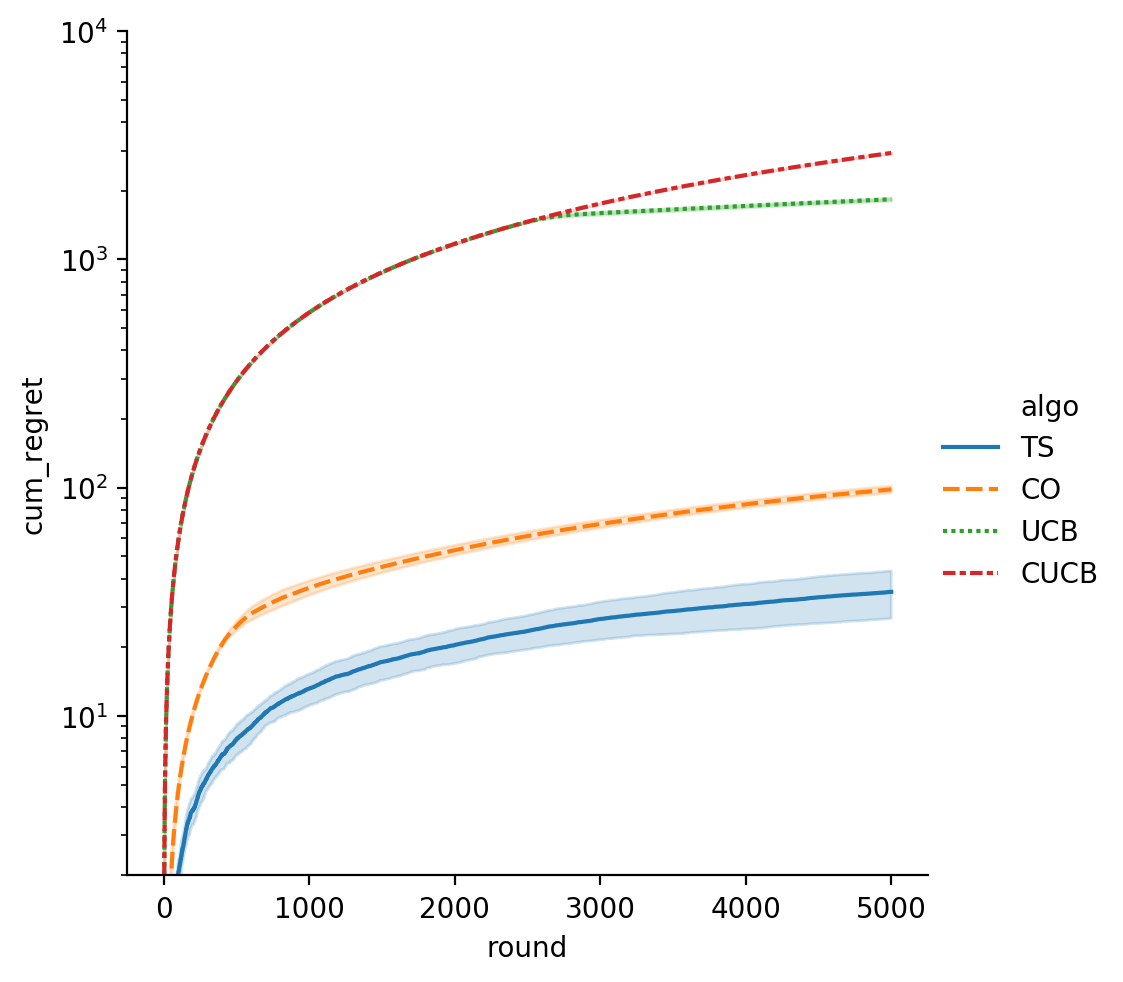}
    \caption{25 nodes, perfect linear generalization.}
    \label{fig:25-node}
\end{figure}

\begin{figure}
    \centering
    \includegraphics[width=0.8\linewidth]{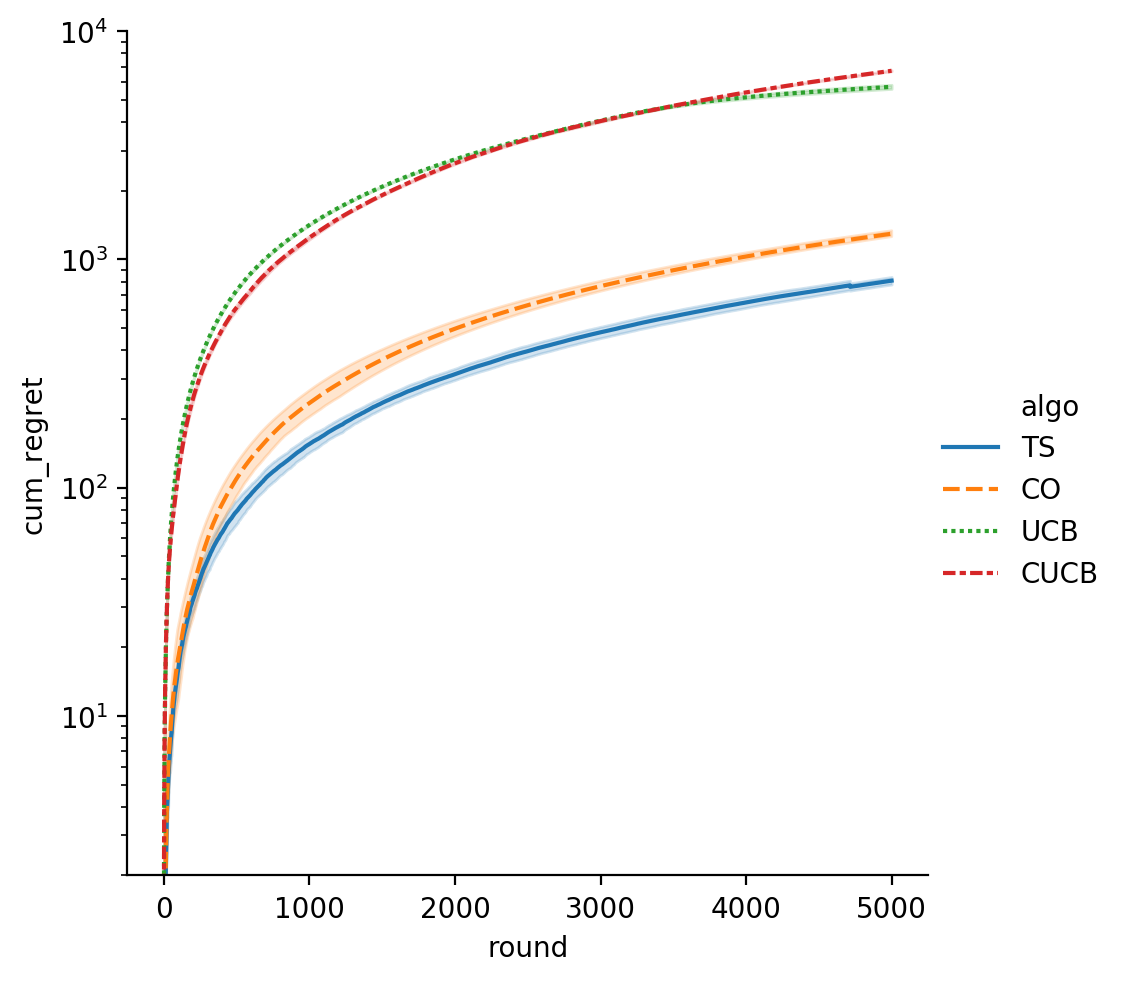}
    \caption{50 nodes, random edge weights.}
    \label{fig:50-node}
\end{figure}

Since the optimal oracle for IMB is NP-hard to find, we are not able to directly compute the cumulative scaled regret. Instead, we report a proxy for it. In Appendix \ref{apd:proxy}, we formally define the proxy and show that in expectation, it upper bounds $\eta$ times the true cumulative scaled regret. We average the proxy over 5 realizations for each online learning algorithm to produce the lines in Figure \ref{fig:25-node} and \ref{fig:50-node}. The shades show the standard deviation of the proxy at each round (plotted with {\tt{seaborn.regplot(..., 'ci'='sd',...)}} in Python). 

As we can see, CO and TS outperform the UCB-based algorithms by a large margin in both network instances. CO performs slightly worse than TS but is much better than UCB and CUCB. Also, with or without perfect linear generalization of edge weights, algorithms assuming linear generalization (i.e., CO, TS, UCB) in general outperform the one that does not (CUCB). 


\bibliography{source.bib}
\newpage
\appendix
	\section{Regret analysis of Algorithm \ref{alg:TS_IMcB}} \label{apd:online}
	\subsection{Definitions}
We use $n = |\cV|$ and $m = |\cE|$ to denote the number of nodes and edges in the given network $\cG = (\cV,\cE)$ respectively. We use $T$ to denote the number of rounds in online learning. For any two edge weights functions $u: \cE \mapsto [0,1]$ and $w: \cE \mapsto [0,1]$, if $u(e) \geq w(e) \; \forall e\in \cE$, then we write $u \geq w$. 
	
Recall that for any seed set $S \subseteq \cV$ and edge weights $w:\cE \mapsto [0,1]$, $f(S,w)$ denotes the expected number of nodes activated during an IC diffusion initiated from $S$ under $w$; the expectation is over the randomness of edge realizations under edge weights $w$. Now let $f(S,w, v)$ be the probability that node $v$ is activated given seed set $S$ and edge weights $w$ during the diffusion. Then by definition, $$f(S,w) = \sum\limits_{v\in \cV} f(S,w,v).$$

We also defined the scaled regret in round $t$ as $R^\eta_t := f(S^*,\mathbf{w}_t) - \frac{1}{\eta}f(S_t, \mathbf{w}_t)$, where $S_t$ is the seed set chosen by Algorithm \ref{alg:TS_IMcB} in round $t$, and $\mathbf{w}_t$ encodes the edge realizations of the IC diffusion in round $t$ under the true edge weights function $\bar{w}:\cE \mapsto [0,1]$. 

Now define $\cH_t$ to be the $\sigma$-algebra generated by the observed edge realizations up to the end of round $t$, with $\cH_0 = \emptyset$. By definition, $\theta_t$ and $\mathbf{M}_{t-1}$ are $\cH_{t-1}$-measurable. Further define $\cF_t$ as the $\sigma$-algebra generated by the observed edge realizations up to the end of round $t-1$ as well as $\tilde \theta_{\tau}$ for $\tau \leq t$, with $\cF_0 = \emptyset$. Besides $\theta_t$ and $\mathbf{M}_{t-1}$, $\tilde w_t$ and $\tilde u_t$ are also $\cF_t$-measurable.

Recall that for $t \geq 1$, $$\alpha_t = \sqrt{d \ln (1+  \frac{tm}{ d} ) + 4\ln t} + D,$$ where $D$ is a known upper bound for $\|\theta^*\|_2$. 

For $t \geq 1$, let $$\beta_t = v \alpha_t (\sqrt{2\ln 2t } +\sqrt{2\ln m + 4\ln t}).$$ 

Further, for any real vector $y \in \mathbb{R}^d$ and positive semi-definite matrix $\mathbf{M} \in \mathbb{R}^{d\times d}$ let $\|y\|_\mathbf{M} = \sqrt{y^\top \mathbf{M} y}$ be a norm of $y$ weighted by $\mathbf{M}$.

For each round $t$, we define two favorable events $\xi_t$ and $\delta_t$, where 
$$\xi_t := \{|x_e^\top\theta_t - x_e^\top\theta^*| \leq \alpha_t \|\mathbf{x}_e\|_{\mathbf{M}_{t-1}^{-1}} \; \forall e\in \cE\},$$
$$\delta_t := \{|\tilde w_t(e) - x_e^\top\theta_t| \leq \beta_t \|\mathbf{x}_e\|_{\mathbf{M}_{t-1}^{-1}} \; \forall e\in \cE\}.$$
We use $\bar{\xi}_t$ and $\bar{\delta}_t$ to denote the complements of these two events. Also note that both $\xi_t$ and $\delta_t$ are $\cF_t$-measurable, and $\xi_t$ is $\cH_{t-1}$-measurable.

Define $O(e,S,w)$ as the event that edge $e = (a,b)$'s realization is observed by the agent during a diffusion under seed set $S$ and edge weights $w$. Recall that $O(e,S,w)$ happens if an only if $a$ is activated during the diffusion. We use $O_t(e,S,w)$ to denote this event for the diffusion in round $t$.
Let $S_t$ be the seed set chosen by Algorithm \ref{alg:TS_IMcB} in round $t$.
We use $O_t(e)$ as an abbreviation for $O_t(e,S_t,\bar{w})$. Since $\bar{w}$ is the true edge weights and $S_t$ is the seed set that is actually chosen for round $t$, the $O_t(e)$'s are events that the agent can actually observe.

Use $\ind\{\cdot\}$ to denote the indicator function. Namely, for any event $A$, $\ind\{A\} = 1$ if $A$ happens, and $0$ otherwise.

\subsubsection{Maximum observed relevance.} \label{apd:mor}
We define a crucial network-dependent complexity metric $C_*$ which was originally proposed in \cite{wen2017online}. Here we extend its definition to our budgeted settings. 
First recall that $\cE^o_t$ is the set of edges whose realizations are observed in round $t$.  Let $P_{S, e}= \PP\{e \in \cE^o_t | S\}$, i.e., the probability that edge $e$'s realization is observed in round $t$ given seed set $S$. In the IC model, $P_{S, e}$ purely depends on the network topology and the true edge weights $\bar{w}$. We say an edge $e \in \cE$ is \textit{relevant} to a node $v \in V\backslash S$ if there exists a path $P$  that starts from any influencer $s \in S$ and ends at $v$, such that $e \in P$ and $P$ contains only one influencer, namely the starting node $s$. Use $\cE_{S, v}$ to denote the set of edges relevant to node $v$ with respect to $S$. Note that $\cE_{S, v}$ depends only on the topology of the network. Similarly, from each edge's perspective, define $N_{S,e} := \sum_{v \in \mathcal{V}\backslash S}\mathbf{1}\{e\text{ is relevant to } v \text{ under } S\}$, i.e., number of non-seed nodes that $e$ is relevant to with respect to $S$. $N_{S,e}$ also depends only on the network topology. With this notation, we define $$C_*:=\max_{S \subseteq \mathcal{V}} \sqrt{\sum_{e\in \cE}N_{S,e}^2 \, P_{S, e}}.$$ Clearly, $C_*$ depends only on the network topology and edge weights. Also, it is upper bounded by $n\sqrt{m}$. $C_*$ is referred to as the \textit{maximum observed relevance} in \cite{wen2017online}.	

\subsection{Preliminaries \& concentration results}

We first provide the following lemma to bound the difference in activation probabilities of node $v$ under two different edge weights functions.
\begin{lemma} 
\label{lemma:diff}
Let $w: \cE \mapsto [0,1]$ and $u: \cE \mapsto [0,1]$ be two edge weights functions. For any seed set $S \subseteq \cV$ and node $v\in \cV$, we have 
\begin{align*}
&|f(S, u, v) - f(S,w,v) |\\ &\leq \sum_{e\in\cE_{S,v}} \EE[\mathds{1}\{O(e,S, w)\}] | u(e) - w(e) |,
\end{align*}
where $\cE_{S,v}$ is the collection of edges relevant to $v$ under $S$ (see definition in Section \ref{apd:mor}). The expectation is over the randomness of the edge realizations under edge weights $w$.
\end{lemma}

\textit{Proof.}
Let $l(e) = \min(u(e),w(e))$, then we have 
\begin{align*}
&|f(S, u, v) - f(S,w,v)|\\ 
= & \Big | f(S, u, v) - f(S, l, v)+ f(S, l, v) - f(S,w,v)  \Big |\\
\leq & \Big | f(S, u, v) - f(S, l, v)| + |f(S, l, v) - f(S,w,v)  \Big | \\
\leq & \sum_{e\in\cE_{S,v}} \EE[\mathds{1}\{O(e,S,l)\}] \Big (|u(e) -  l(e)| \Big ) \\
&+ \sum_{e\in\cE_{S,v}} \EE[\mathds{1}\{O(e,S,l)\}]  \Big (| l(e) - w(e) |\Big )
\\
\leq & \sum_{e\in\cE_{S,v}} \EE[\mathds{1}\{O(e,S,w)\}] \Big (|u(e) -  l(e)| \Big ) \\
&+ \sum_{e\in\cE_{S,v}} \EE[\mathds{1}\{O(e,S,w)\}]  \Big (| l(e) - w(e) |\Big )\\
= & \sum_{e\in\cE_{S,v}} \EE[\mathds{1}\{O(e,S,w)\}] | u(e) - w(e) |,
\end{align*}
where the second inequality follows from Theorem 3 in \cite{wen2017online},
the third inequality is due to the fact that $\EE[\mathds{1}\{O(e,S,l)\}] \leq \EE[\mathds{1}\{O(e,S,w)\}]$ for $l \leq w$, and the last equality comes from the fact that $l(e) = \min \Big (w(e),u(e) \Big )$.
\qed
%%%%%%%%%%%%%%
%%%%%%%%%%%%%%%%

We now present a lemma that connects $\tilde w_t(e)$'s with the maximum of $t$ standard normal random variables. 

\begin{lemma} \label{lemma:dist_w}
In round $t$, let $Z_j$ for $j=1,\cdots,t$ be $t$ iid standard normal random variables. For any $s \in \mathbb{R}$ and $\cH_{t-1}$, we have  
$$\PP \Big ( \frac{\tilde w_t(e) - \mathbf{x}_e^\top \theta_t}{ \alpha_t \|x_e \|_{\mathbf{M}_{t-1}^{-1}}  } \leq s \; \Big | \cH_{t-1} \Big ) = \PP \Big(\max_{j=1,\cdots,t} Z_j\leq s /v \Big).$$
\end{lemma}
\textit{Proof.}
We prove the lemma by induction. As $\tilde w_0(e) := -\infty$ 
for all $e\in \cE$, it is easy to see that $\tilde w_1(e) = \mathbf{x}_e^\top \tilde \theta_1 \; \forall e\in \cE$, where $\tilde{\theta}_{1}$ is sampled from a multivariate Gaussian distribution with mean vector $\theta_{1}$ and covariance matrix $v^2 \alpha_1^2\mathbf{M}_{0}^{-1}$. 
Therefore, for each $e\in \cE$, $\tilde w_1(e)$ has mean $x_e^\top   \theta_1$ and standard deviation 
$v\alpha_1 \|\mathbf{x}_e \|_{\mathbf{M}_{0}^{-1}}$. We thus have 
\begin{align*}
    \PP \Big ( \frac{\tilde w_1(e) - \mathbf{x}_e^\top \theta_1}{ \alpha_1 \|\mathbf{x}_e \|_{\mathbf{M}_{0}^{-1}}  } \leq s \; \Big | \cH_0\Big ) &=   \PP \Big(\frac{ \mathbf{x}_e^\top \tilde \theta_1 - \mathbf{x}_e^\top \theta_1}{ \alpha_1 \|\mathbf{x}_e \|_{\mathbf{M}_{0}^{-1}}  } \leq s  \Big)\\ 
    &= \PP \Big( Z_1\leq s/v \Big),
\end{align*}
which implies the correctness of the lemma statement for $t=1$. 

Suppose the statement is true for round $t-1$.  In round $t$, by definition, 
$\tilde{\theta}_{t}$ is sampled from a multivariate Gaussian distribution with mean vector $\theta_{t}$ and covariance matrix $v^2\alpha_t^2\mathbf{M}_{t-1}^{-1}$. 
Therefore, for each $e\in \cE$, $\mathbf{x}_e^\top \tilde \theta_t$ has mean $x_e^\top   \theta_t$ and standard deviation 
$v\alpha_t \|\mathbf{x}_e \|_{\mathbf{M}_{t-1}^{-1}}$.
Further recall that
\begin{align*}
\tilde w_t(e) := \max \Big (&\mathbf{x}_e^\top \theta_t+\frac{\tilde w_{t-1}(e) - \mathbf{x}_e^\top  \theta_{t-1}}{\alpha_{t-1}\|\mathbf{x}_e\|_{\mathbf{M}_{t-2}^{-1}}} \alpha_t \|\mathbf{x}_e \|_{\mathbf{M}_{t-1}^{-1}}, \\ &\mathbf{x}_e^\top \tilde \theta_t \Big ).
\end{align*}
Thus, we have 
\begin{align*}
&\PP \Big ( \frac{\tilde w_t(e) - \mathbf{x}_e^\top \theta_t}{ \alpha_t \|\mathbf{x}_e \|_{\mathbf{M}_{t-1}^{-1}}  } \leq s \; \Big | \cH_{t-1}\Big ) \\
= & \PP \Big (\max \Big (\frac{\tilde w_{t-1}(e) - \mathbf{x}_e^\top  \theta_{t-1}}{\alpha_{t-1} \|\mathbf{x}_e\|_{\mathbf{M}_{t-2}^{-1}}} , \frac{x_e^\top \tilde \theta_t  - \mathbf{x}_e^\top \theta_t}{\alpha_t \|\mathbf{x}_e \|_{\mathbf{M}_{t-1}^{-1}} }\Big ) \leq s \; \Big | \cH_{t-1}\Big )
\\
= & \PP \Big (\frac{\tilde w_{t-1}(e) - \mathbf{x}_e^\top  \theta_{t-1}}{\alpha_{t-1} \|\mathbf{x}_e\|_{\mathbf{M}_{t-2}^{-1}}} \leq s,  \frac{x_e^\top \tilde \theta_t  - \mathbf{x}_e^\top \theta_t}{\alpha_t \|\mathbf{x}_e \|_{\mathbf{M}_{t-1}^{-1}} } \leq s \; \Big | \cH_{t-1}\Big )
\\
= & \PP \Big (\frac{\tilde w_{t-1}(e) - \mathbf{x}_e^\top  \theta_{t-1}}{\alpha_{t-1} \|\mathbf{x}_e\|_{\mathbf{M}_{t-2}^{-1}}} \leq s\; \Big | \cH_{t-1}\Big ) \\ \cdot & \PP \Big (\frac{x_e^\top \tilde \theta_t  - \mathbf{x}_e^\top \theta_t}{\alpha_t \|\mathbf{x}_e \|_{\mathbf{M}_{t-1}^{-1}} } \leq s \; \Big | \cH_{t-1}\Big )
\\
=& \PP\Big (\max_{i=1,\cdots,t-1}Z_i \leq s/v\Big )  \PP\Big (Z_t \leq s/v\Big ) \\
= & 
\PP \Big(\max_{i=1,\cdots,t} Z_i\leq s /v\Big).
\end{align*}
The third equality above follows from the independence of $\tilde \theta_t$ and $\tilde w_{t-1}$ given $\cH_{t-1}$. The fourth equality is by the inductive hypothesis and the property of Gaussian random variables.
\qed
%%%%%%%%%%%%%%
%%%%%%%%%%%

We now provide two concentration results for $\theta_t$ and $\tilde w_t$ respectively.
\begin{lemma}[Concentration of $\theta_t$]\label{lemma:concentration_theta}
For every $t \geq 1$,
$$\PP(\xi_t) = \PP\Big (|\mathbf{x}_e^\top \theta_t - \bar{w}(e)| \leq \alpha_t \|\mathbf{x}_e\|_{\mathbf{M}_{t-1}^{-1}} \; \forall e\in \cE \Big) \geq 1-\frac{1}{t^2}.$$
\end{lemma}
\textit{Proof.}
This result is implied by Lemma 2 in \cite{wen2017online}.
\qed
%%%%%%%%%%%%%%%%%
%%%%%%%%%%%%%%%%%
\begin{lemma}[Concentration of $\tilde w_t$] \label{lemma:concentration_wtilde}
For all $e \in \cE$, 
\begin{align*}
&\PP(\delta_t \; | \cH_{t-1})\\ 
&= \PP\Big(|\tilde w_t(e) - \mathbf{x}_e^\top \theta_t | \leq \beta_t \|\mathbf{x}_e\|_{\mathbf{M}_{t-1}^{-1}} \; \forall e\in \cE \; \Big | \cH_{t-1}\Big)\\ 
&\geq 1 - \frac{1}{t^2}.
\end{align*}
\end{lemma}
\textit{Proof.}
From Lemma \ref{lemma:dist_w}, we have that for any $s\in \mathbb{R}$,
$$\PP \Big ( \frac{\tilde w_t(e) - \mathbf{x}_e^\top \theta_t}{ \alpha_t \|x_e \|_{\mathbf{M}_{t-1}^{-1}}  } \leq s \; \Big | \cH_{t-1} \Big ) = \PP \Big(\max_{j=1,\cdots,t} Z_j\leq s/v\Big),$$ where the $Z_j$'s are iid standard normal random variables.
Therefore, we have 
\begin{align*}
&\PP \Big ( \frac{|\tilde w_t(e) - \mathbf{x}_e^\top \theta_t|}{ \alpha_t \|x_e \|_{\mathbf{M}_{t-1}^{-1}}  } \leq s \; \Big | \cH_{t-1} \Big )\\ &= \PP \Big(|\max_{j=1,\cdots,t} Z_j|\leq s/v \Big)\\ 
&\geq \PP \Big(\max_{j=1,\cdots,t} |Z_j|\leq s/v \Big).
\end{align*}

Set $s = v( \sqrt{2\ln 2t } +\sqrt{2\ln m + 4\ln t}) $. Since $\beta_t = s\alpha_t$ by definition, we have 
\begin{align*}
&\PP \Big ( \frac{|\tilde w_t(e) - \mathbf{x}_e^\top \theta_t|}{ \alpha_t \|x_e \|_{\mathbf{M}_{t-1}^{-1}}  } \leq s \; \Big | \cH_{t-1} \Big )\\ 
&= \PP\Big(|\tilde w_t(e) - \mathbf{x}_e^\top \theta_t | \leq \beta_t \|\mathbf{x}_e\|_{\mathbf{M}_{t-1}^{-1}}| \cH_{t-1}\Big).
\end{align*}

By Lemma \ref{lemma:max_normal} (set $\delta = 1/mt^2$), we have 
$$\PP \Big(\max_{j=1,\cdots,t} |Z_j|\leq s/v \Big) \geq 1-\frac{1}{mt^2}.$$

We have thus established that

\begin{align*}
&\PP\Big(|\tilde w_t(e) - \mathbf{x}_e^\top \theta_t | \leq \beta_t \|\mathbf{x}_e\|_{\mathbf{M}_{t-1}^{-1}} \big | \cH_{t-1}\Big)\\ 
&\geq \PP \Big(\max_{j=1,\cdots,t} |Z_j|\leq s/v \Big) \geq 1-\frac{1}{mt^2}.
\end{align*}

Now by union of probability, we have the desired
$$\PP\Big(|\tilde w_t(e) - \mathbf{x}_e^\top \theta_t | \leq \beta_t \|\mathbf{x}_e\|_{\mathbf{M}_{t-1}^{-1}} \; \forall e\in \cE \; \Big | \cH_{t-1}\Big) \geq 1 - \frac{1}{t^2}.$$
\qed
%%%%%%%%%%%%%%
%%%%%%%%%%%%%%
%%%%%%%%%%%%%%%
\subsection{Proof of Theorem \ref{thmTS-CO}}
\textit{Proof of Theorem \ref{thmTS-CO}:}

The scaled regret in round $t$ is $R^\eta_t := f(S^*,\mathbf{w}_t) - \frac{1}{\eta}f(S_t, \mathbf{w}_t)$. $\mathbf{w}_t$ encodes random edge realizations under true edge weights $\bar{w}$. $S_t$ is the seed set chosen by Algorithm \ref{alg:TS_IMcB}in round $t$.
The expected round scaled regret given $\cF_t$ is $\E(R^\eta_t|\cF_t) := \E\Big(f(S^*,\bar{w}) - \frac{1}{\eta}f(S_t, \bar{w}) \Big |\cF_t\Big)$, where the expectation is over the randomness of $S_t$ (since our oracle is a randomized oracle) and $S^*$ (since the optimal oracle can also be randomized).

Since $\xi_t$ is $\cF_t$-measurable, we can decompose $\E(R^\eta_t)$ as follows:
\begin{equation}\label{eq:decomp}
\begin{split}
    \E(R_t^{\eta}) &= \E(R_t^\eta|\xi_t)\PP(\xi_t) + \E(R_t^\eta|\bar{\xi}_t)\PP(\bar{\xi}_t)\\
    &\leq \underbrace{\E(R_t^\eta|\xi_t)}_{Q_0}\PP(\xi_t) + \frac{n}{t^2}
\end{split}
\end{equation}
where the first inequality follows because 1) $R_t^\eta$ is upper bounded by $n$ and 2) $\PP(\bar{\xi}_t) \leq 1/t^2$ by Lemma \ref{lemma:concentration_theta}.

Next, we decompose $Q_0$.
\begin{equation} \label{eq:decomp_1}
\begin{split}
    Q_0
    = &\EE \Big[f(S^*,\bar w) - \frac{1}{\eta}f(S_t, \tilde u_t)\\ &+ \frac{1}{\eta}f(S_t, \tilde u_t) -  \frac{1}{\eta}f(S_t, \bar w) \Big | \xi_t \Big]\\
    = &\underbrace{\EE \Big[ f(S^*,\bar w) - \frac{1}{\eta}f(S_t, \tilde u_t)\Big | \xi_t \Big]}_{Q_1}\\ 
    & + \frac{1}{\eta} \underbrace{\EE \Big[f(S_t, \tilde u_t) - f(S_t, \bar w)\Big |\xi_t \Big].}_{Q_2}
    \end{split}
\end{equation}
We upper bound $Q_1$ first.

Use $S^*(\tilde u_t)$ to denote the (possibly randomized) optimal seed set given budget $B/T$ under edge weights $\tilde u_t$. By Theorem \ref{offline_thm2}, we have for any $\cF_t$
\begin{align*}
\frac{1}{\eta}\E(f(S_t,\tilde u_t)|\cF_t) &\geq \E(f(S^*(\tilde u_t),\tilde u_t)|\cF_t)\\ &\geq \E(f(S^*,\tilde u_t)|\cF_t).
\end{align*}
Therefore, we have
\begin{equation}
\begin{split}
Q_1 &= \EE \Big[ f(S^*,\bar w) - \frac{1}{\eta} f(S_t, \tilde u_t)\Big | \xi_t \Big]\\
& \leq \EE \Big[ f(S^*,\bar w) - f(S^*, \tilde u_t)\Big |\xi_t \Big]. 
\end{split}
\end{equation}
We make the following observations: 1) $f(S^*,\bar w) - f(S^*, \tilde u_t) \leq n$, 2) if $\tilde u_t \geq \bar{w}$, then $f(S^*,\bar w) - f(S^*, \tilde u_t) \leq 0$. We therefore have the upper bound
$$\E\Big[f(S^*,\bar w) - f(S^*, \tilde u_t) \Big|\xi_t\Big] \leq   n \big (1-\PP(\tilde u_t \geq \bar w |\xi_t) \big ).
$$
The above leads to the following upper bound for $Q_1$: 
\begin{equation} \label{eq:Q1_prob}
Q_1 \leq n \big (1-\PP(\tilde u_t \geq \bar w \;| \xi_t) \big )= n \big (1-\PP(\tilde w_t \geq \bar w \;| \xi_t) \big ),
\end{equation}
where the equality follows because for each $e\in \cE$, $0 \leq \bar{w}(e) \leq 1$ and $\tilde u_t(e)$ is the projection of $\tilde w_t(e)$ on $[0,1]$ for each $e \in \cE$.

Now we lower bound $\PP(\tilde w_t \geq \bar w \;| \xi_t)$.
\begin{equation} \label{eq:Q1_condition_first}
\begin{split}
&\PP(\tilde w_t \geq \bar w \;| \xi_t)\\ &= \E(\PP(\tilde w_t \geq \bar{w}\;|\cH_{t-1})\;|\xi_t)\\ 
&= \E(\PP(\tilde w_t(e) \geq \bar{w}(e) \; \forall e\in \cE\;|\cH_{t-1})\;|\xi_t)\\ 
&= \E\Big[\PP \Big ( \frac{\tilde w_t(e) - \mathbf{x}_e^\top \theta_t}{\alpha_t \|\mathbf{x}_e\|_{\mathbf{M}_{t-1}^{-1}}} \geq  \frac{\bar w(e) - \mathbf{x}_e^\top \theta_t}{\alpha_t \|\mathbf{x}_e\|_{\mathbf{M}_{t-1}^{-1}}} \; \forall e\in \cE \Big | \cH_{t-1} \Big ) \; \Big | \xi_t \Big]\\
&\geq \E\Big[\PP \Big ( \frac{\tilde w_t(e) - \mathbf{x}_e^\top \theta_t}{\alpha_t \|\mathbf{x}_e\|_{\mathbf{M}_{t-1}^{-1}}} \geq 1 \; \forall e\in \cE \Big | \cH_{t-1} \Big ) \; \Big | \xi_t \Big]\\
&= 1- \E\Big[\PP \Big ( \frac{\tilde w_t(e) - \mathbf{x}_e^\top \theta_t}{\alpha_t \|\mathbf{x}_e\|_{\mathbf{M}_{t-1}^{-1}}} \leq 1 \; \exists e\in \cE \Big | \cH_{t-1} \Big ) \; \Big | \xi_t \Big]\\
& \geq 1- \sum_{e\in \cE} \E\Big[\PP \Big ( \frac{\tilde w_t(e) - \mathbf{x}_e^\top \theta_t}{\alpha_t \|\mathbf{x}_e\|_{\mathbf{M}_{t-1}^{-1}}} \leq 1 \;\Big | \cH_{t-1} \Big ) \; \Big | \xi_t \Big]\\
&=1- m \PP \Big ( \max_{j=1,\cdots,t}Z_j \leq  1/v\Big ) = 1- m \PP(Z_j \leq 1/v)^t\\
& \geq 1- m \Big ( 1-\frac{v}{4\sqrt{\pi } } e^{-1/2v^2} \Big )^t.
\end{split}
\end{equation}
Above, the first equality follows because $\xi_t$ is $\cH_{t-1}$-measurable; the first inequality follows because under $\xi_t$, we have 
$|x_e^\top\theta_t - \bar{w}(e)| \leq \alpha_t \|\mathbf{x}_e\|_{\mathbf{M}_{t-1}^{-1}} \; \forall e\in \cE$; the second last line follows by Lemma \ref{lemma:dist_w} (the $Z_j$'s are iid standard normal random variables); the last inequality follows from Lemma \ref{lemma:Abramowitz}.

Combining \eqref{eq:Q1_prob} and \eqref{eq:Q1_condition_first}, we have 
\begin{equation} \label{eq:decomp_2}
Q_1 \leq nm\Big ( 1-\frac{v}{4\sqrt{\pi } } e^{-1/2v^2} \Big )^t.
\end{equation}

We now bound $Q_2 = \EE \Big[f(S_t, \tilde u_t) - f(S_t, \bar w) \Big |\xi_t \Big]$.

We can decompose $Q_2$ by conditioning on $\delta_t$ and $\bar{\delta}_t$. Namely,
\begin{equation} \label{eq:Q2decompQ3}
\begin{split}
&Q_2 = \EE \Big[f(S_t, \tilde u_t) - f(S_t, \bar w) \Big |\xi_t \Big]\\
= &\EE \Big[f(S_t, \tilde u_t) - f(S_t, \bar w) \Big |\delta_t,\xi_t \Big]\PP(\delta_t|\xi_t)\\
&+ \EE \Big[f(S_t, \tilde u_t) - f(S_t, \bar w) \Big |\bar{\delta}_t,\xi_t \Big]\PP(\bar{\delta}_t|\xi_t)\\
\leq &\EE \Big[f(S_t, \tilde u_t) - f(S_t, \bar w) \Big |\delta_t,\xi_t \Big]\PP(\delta_t|\xi_t)\\ &+ n \E(\PP(\bar{\delta}_t|\cH_{t-1})|\xi_t)\\
\leq &\underbrace{\EE \Big[f(S_t, \tilde u_t) - f(S_t, \bar w) \Big |\delta_t,\xi_t \Big]}_{Q_3}\PP(\delta_t|\xi_t) + \frac{n}{t^2}.
\end{split}
\end{equation}
The first inequality follows because 1)$f(S_t, \tilde u_t) - f(S_t, \bar w) \leq n$ and 2) $\xi_t$ is $\cH_{t-1}$-measurable; the second inequality follows from Lemma \ref{lemma:concentration_wtilde}.

We now upper bound $Q_3$.

First, we decompose $f(S_t, \tilde u_t) - f(S_t, \bar w)$ into a sum over individual node activation probabilities and upper bound it using the property of absolute values. Namely, 
\begin{equation} \label{eq:Q3_decompf_0}
\begin{split}
&f(S_t, \tilde u_t) - f(S_t, \bar w)\\ &= \sum_{v\in \cV \setminus S_t} f(S_t, \tilde u_t, v) - f(S_t, \bar w, v)\\
& \leq \sum_{v\in \cV \setminus S_t} |f(S_t, \tilde u_t, v) - f(S_t, \bar w, v)|.
\end{split}
\end{equation}

Now using Lemma \ref{lemma:diff}, we have 
\begin{equation} \label{eq:Q3_decompf_1}
\begin{split}
&\sum_{v\in \cV \setminus S_t} |f(S_t, \tilde u_t, v) - f(S_t, \bar w, v)|\\ &\leq \sum_{v\in \cV \setminus S_t}\sum_{e\in\cE_{S_t,v}} \EE[\mathds{1}\{O_t(e,S_t, \bar{w})\}] | \tilde u_t(e) - \bar{w}(e) |\\
& = \sum_{v\in \cV \setminus S_t}\sum_{e\in\cE_{S_t,v}} \EE[\mathds{1}\{O_t(e)\}] | \tilde u_t(e) - \bar{w}(e) |.
\end{split}
\end{equation}
The expectation above is over the randomness of edge realizations of the diffusion process in round $t$ under the true edge weights $\bar{w}$.

From \eqref{eq:Q3_decompf_0} and \eqref{eq:Q3_decompf_1}, we have 
\begin{equation} \label{eq:Q3_decompf_2}
\begin{split}
&Q_3 = \EE \Big[f(S_t, \tilde u_t) - f(S_t, \bar w) \Big | \delta_t, \xi_t \Big]\\
\leq &\EE \Big[\sum_{v\in \cV \setminus S_t}\sum_{e\in\cE_{S_t,v}} \mathds{1}\{O_t(e)\} |\tilde u_t(e) - \bar{w}(e) \Big | \delta_t, \xi_t \Big].
\end{split}
\end{equation}
Note the inner expectation above is over both the randomness of $S_t$ (as it is output by a randomized seeding oracle) and over the edge realizations in round $t$.

We now examine $|\tilde u_t(e) - \bar w(e)|$ for each $e \in \cE$ given $\xi_t$ and $\delta_t$. Recall that $\tilde u_t(e) = \text{Proj}_{[0,1]} \tilde w_t(e)$. Also, since $\bar w(e)$ is the true edge weight on $e$, we have $0 \leq \bar w(e) \leq 1$. Therefore, if $\tilde w_t(e)>1$ or $\tilde w_t(e)<0$, we clearly have $|\tilde u_t(e) - \bar w(e)| \leq |\tilde w_t(e) - \bar w(e)|$. On the other hand, if $0 \leq \tilde w_t(e) \leq 1$, then $\tilde u_t(e) = \tilde w_t(e)$, and we have $|\tilde u_t(e) - \bar w(e)| = |\tilde w_t(e) - \bar w(e)|$. As a result, for any $e \in \cE$, 
\begin{equation} \label{eq:Q2_decompf_3}
\begin{split}
&|\tilde u_t(e) - \bar w(e)| \leq |\tilde w_t(e) - \bar w(e)|\\ & = |\tilde w_t(e) - x_e^\top  \theta_t + x_e^\top  \theta_t - \bar w(e)| \\
& \leq |\tilde w_t(e) - x_e^\top \theta_t| + |x_e^\top \theta_t - \bar w(e)|\\
& \leq (\alpha_t + \beta_t)\|\mathbf{x}_e\|_{\mathbf{M}_{t-1}^{-1}},
\end{split}
\end{equation}
where the last inequality is due to the definition of $\delta_t$ and $\xi_t$.

From \eqref{eq:Q3_decompf_2} and \eqref{eq:Q2_decompf_3}, we have 
\begin{equation} \label{eq:decomp_3}
\begin{split}
Q_3 \leq &(\alpha_t + \beta_t) \\
& \cdot \EE \Big[\sum_{v\in \cV \setminus S_t}\sum_{e\in\cE_{S_t,v}} \mathds{1}\{O_t(e)\} \|\mathbf{x}_e\|_{\mathbf{M}_{t-1}^{-1}} \Big)\Big | \delta_t, \xi_t \Big].
\end{split}
\end{equation}

From the above inequalities, we have 
\begin{equation} \label{eq:secondlast}
\begin{split}
\E(R_t^{\eta}) 
\leq &Q_0\PP(\xi_t) + \frac{n}{t^2}\\
= &(Q_1 + \frac{Q_2}{\eta})\PP(\xi_t) + \frac{n}{t^2}\\
\leq & (nm   \Big ( 1-\frac{v}{4\sqrt{\pi } } e^{-1/2v^2} \Big )^t\\  
&+ \frac{Q_3\PP(\delta_t|\xi_t) + n/t^2}{\eta})\PP(\xi_t) + \frac{n}{t^2}\\
\leq &nm   \Big ( 1-\frac{v}{4\sqrt{\pi } } e^{-1/2v^2} \Big )^t  +
  \frac{Q_3}{\eta}\PP(\delta_t,\xi_t) + \frac{2n}{t^2}\\
\leq &nm    \Big ( 1-\frac{v}{4\sqrt{\pi } } e^{-1/2v^2} \Big )^t +\frac{2n}{t^2}\\
& + \frac{\alpha_t+\beta_t}{\eta} \cdot \PP(\delta_t, \xi_t)\\ 
&\;\; \cdot \EE \Big[\sum_{v\in \cV \setminus S_t}\sum_{e\in\cE_{S_t,v}} \mathds{1}\{O_t(e)\}\|\mathbf{x}_e\|_{\mathbf{M}_{t-1}^{-1}}\Big | \delta_t, \xi_t \Big]\\
\leq &nm    \Big ( 1-\frac{v}{4\sqrt{\pi } } e^{-1/2v^2} \Big )^t    +\frac{2n}{t^2}\\ 
& + \frac{\alpha_t+\beta_t}{\eta}\\ 
&\;\;\cdot \EE \Big[\sum_{v\in \cV \setminus S_t}\sum_{e\in\cE_{S_t,v}} \mathds{1}\{O_t(e)\}\|\mathbf{x}_e\|_{\mathbf{M}_{t-1}^{-1}} \Big].
\end{split}    
\end{equation}
More specifically, the first inequality above is by \eqref{eq:decomp}, 
the first equality is by
\eqref{eq:decomp_1}, 
the second inequality is by \eqref{eq:decomp_2} and \eqref{eq:Q2decompQ3},
the third inequality follows because $\PP(\delta_t|\xi_t)\PP(\xi_t) = \PP(\delta_t,\xi_t)$ and $\PP(\xi_t)\leq 1$,
the fourth inequality is by
\eqref{eq:decomp_3},
the last inequality follows because for all $\cF_t$, $$\EE\Big(\sum_{v\in \cV \setminus S_t}\sum_{e\in\cE_{S_t,v}} \mathds{1}\{O_t(e)\}\|\mathbf{x}_e\|_{\mathbf{M}_{t-1}^{-1}} \Big | \cF_t \Big) \geq 0.$$  

The expectation in the last line of \eqref{eq:secondlast} is over both the $\cF_t$ and the randomness of $S_t$ returned by the randomized seeding oracle. 

Define 
$$\tilde p =   1-\frac{v}{4\sqrt{\pi } } e^{-1/2v^2},\;\;\;  N_{S,e} = \sum_{v\in \cV \setminus S} \mathds{1} \{e \in \cE_{S,v}\},$$ 
we can represent the above upper bound of $\E(R_t^\eta)$ as 
\begin{align*}
\E(R_t^\eta) \leq & nm{\tilde p}^t+\frac{2n}{t^2}\\ &+ \frac{\alpha_t+\beta_t}{\eta} \EE \Big[\sum_{e\in\cE} \mathds{1}\{O_t(e)\} N_{S_t,e}\|\mathbf{x}_e\|_{\mathbf{M}_{t-1}^{-1}} \Big].
\end{align*}

Thus, the expected cumulative regret over $T$ rounds is 
\begin{equation} \label{eq:regret01}
\begin{split}
& R^\eta(T) = \E\Big(\sum_{t = 1}^T R_t^\eta\Big)\\
\leq & nm\sum_{t=1}^T \tilde p^t + 2n\sum_{t=1}^T t^{-2}\\ &+ \frac{\alpha_t+\beta_t}{\eta} \underbrace{\EE \Big[\sum_{t=1}^T\sum_{e\in\cE} \mathds{1}\{O_t(e)\}N_{S_t,e}\|\mathbf{x}_e\|_{\mathbf{M}_{t-1}^{-1}} \Big]}_{Q_4} ,
\end{split}
\end{equation}
where the expectation is over the randomness of the seeding oracle and over $\cF_T$.

We now focus our attention on $Q_4$.

By Lemma 1 in \cite{wen2017online}, the following always holds:
\begin{align*}
    &\sum_{t=1}^T \sum_{v \in \cE} \mathds{1}\{O_t(e)\}N_{S_t,e} \|\mathbf{x}_e\|_{\mathbf{M}_{t-1}^{-1}}\\ &\leq \sqrt{  \sum_{t=1}^T \sum_{e\in \cE}\mathds{1}\{O_t(e)\}N_{S_t,e}^2 } \sqrt{\frac{d m \ln (1+ \frac{Tm}{d})}{\ln 2}}.
\end{align*}

Therefore, 
\begin{equation} \label{eq:regret02}
\begin{split}
&Q_4 \leq \sqrt{\frac{d m \ln (1+ \frac{Tm}{d})}{\ln 2}} \EE \Big[ \sqrt{ \sum_{t=1}^T \sum_{e\in \cE}\mathds{1}\{O_t(e)\}N_{S_t,e}^2}\Big],
\end{split}
\end{equation}
where the expectation is over the randomness of the seeding oracle.

Moreover, by definition of $C^*$, we have that for any $t = 1, \cdots, T$,
\begin{equation} \label{eq:def_C*}
\sum_{e\in \cE}\EE\Big[\mathds{1}\{O_t(e)\}N_{S_t,e}^2\Big] \leq C_*^2.
\end{equation}

By Jensen's inequality and \eqref{eq:def_C*}, we have
\begin{equation} \label{eq:regret03}
\begin{split}
&\EE \Big[ \sqrt{ \sum_{t=1}^T \sum_{e\in \cE}\mathds{1}\{O_t(e)\}N_{S_t,e}^2}\Big]\\
& \leq \sqrt{ \sum_{t=1}^T \sum_{e\in \cE}\EE [\mathds{1}\{O_t(e)\}N_{S_t,e}^2]} \leq C^*\sqrt{T}.
\end{split}
\end{equation}

Combining \eqref{eq:regret01}, \eqref{eq:regret02}, \eqref{eq:regret03}, we have 
\begin{equation} \label{eq:regret04}
\begin{split}
R^\eta(T)
\leq & nm\sum_{t=1}^T \tilde p^t + 2n\sum_{t=1}^T t^{-2}+ \frac{\alpha_t+\beta_t}{\eta}Q_4\\
\leq & nm\sum_{t=1}^T \tilde p^t + 2n\sum_{t=1}^T t^{-2}\\ 
&+ \frac{(\alpha_t+\beta_t)C^*}{\eta}\sqrt{\frac{d m T\ln (1+ \frac{Tm}{d})}{\ln 2}}.
\end{split}
\end{equation}
Further, we have 
$$\sum_{t=1}^T \tilde p^t \leq \frac{1}{1-\tilde p} =  \frac{4\sqrt{\pi }  e^{1/2v^2}}{v}, \;\;\; \sum_{t=1}^T t^{-2} \leq \frac{\pi^2}{6}.$$

Combining these facts with \eqref{eq:regret04}, we have the desired
\begin{align*}
R^\eta(T) \leq  & n\Big(\frac{4m\sqrt{\pi }  e^{1/2v^2}}{v} + \frac{\pi^2}{3} \Big)\\ 
& + \frac{(\alpha_T+\beta_T)C^*}{\eta}\sqrt{\frac{d m T\ln (1+ \frac{Tm}{d})}{\ln 2}}\\
\leq & n\Big(\frac{4m\sqrt{\pi }  e^{1/2v^2}}{v} + \frac{\pi^2}{3} \Big)\\ 
& + \frac{(\alpha_T+\beta_T)mn}{\eta}\sqrt{\frac{d T\ln (1+ \frac{Tm}{d})}{\ln 2}}.
\end{align*}
\qed
\subsection{Auxiliary Lemmas}
\begin{lemma}[ \cite{NIPS2019_8578}]\label{lemma:max_normal}
Let $Z_i \sim N(0,1)$, $i=1,\cdots,n$ be $n$ standard Gaussian random variables. Then we have 
$$
\PP \Big ( \max_i |Z_i| \leq \sqrt{2\ln(2n)} + \sqrt{2\ln \frac{1}{\delta}}\Big ) \geq 1-\delta.
$$
\end{lemma}
	\begin{lemma}[ \cite{abramowitz1972handbook}]\label{lemma:Abramowitz}
	For a Gaussian random variable $Z$ with mean $\mu$ and variance $\sigma^2$, for any $z \geq 1$, 
	$$
	\frac{1}{2\sqrt{\pi} z} e^{-z^2/2} \leq \PP(|Z- \mu| \geq \sigma z) \leq \frac{1}{\sqrt{\pi} z} e^{-z^2/2}.
	$$
	\end{lemma}
	
\section{Cumulative Scaled Regret of Online Learning Algorithms for Lin-IMB-L} \label{apd:regretdecomp}

First observe that IMB can be equivalently formulated as the following linear program (LP1): 
\begin{equation} \label{lp1}
\begin{split}
\max \; &\sum_{S \in \mathcal{P}(\mathcal{V})} p(S) f(S, \bar{w})\\ 
\text{s.t} &\sum_{S \in \mathcal{P}(\mathcal{V})} \mathbf{c}(S) p(S) \leq b ;\\ 
& \sum_{S \in \mathcal{P}(\mathcal{V})} p(S) =1; \\
& p(S) \geq 0 \;\forall \; S \in \mathcal{P}(\mathcal{V}),
\end{split}
\end{equation}
where $\mathcal{P(V)}$ is the power set of the node set $\mathcal{V}$. 
		
We have the following lemma.
		
	\begin{lemma} \label{lemma:optimbbar}
		Consider a Lin-IMB-L instance with input graph $\cD=(\cV,\cE)$, edge weights $\bar{w}$, node costs $\mathbf{c}$, and budget $B$. Let $p^*$ be an optimal solution to LP1 for the corresponding IMB problem with $b = B/T$. For $t = 1, \cdots, T$, let $S^*_t$ be the seed set sampled from $\mathcal{P}(\cV)$ following the probability distribution $p^*$. Then the sequence $\{S^*_t\}_{t=1}^T$ is an optimal solution to the Lin-IMB-L instance.
	\end{lemma}
	
	\textit{Proof.}
	It is easy to see that $\sum_{t=1}^{T} \E(\mathbf{c}(S^*_t)) \leq B$ holds from the budget constraint in LP1. To see that $\sum_{t=1}^T\E(f(S^*_t, \bar{w}))$ is maximized, first observe that any optimal strategy to Lin-IMB-L with budget $B$ and $T$ rounds must assume the following form: in each round $t$, the optimal strategy selects seed set $S$ with probability $p_t(S)$ for all $S \in \mathcal{P}(\mathcal{V})$, such that $$\sum_{S \in \mathcal{P}(\mathcal{V})} p_t(S) = 1 \; \text{ and }\; \sum_{t = 1}^T \sum_{S \in \mathcal{P}(\mathcal{V})} \mathbf{c}(S)p_t(S) \leq B.$$ Furthermore, because it is the optimal strategy, its corresponding expected reward, $$\sum_{t = 1}^T \sum_{S \in \mathcal{P}(\mathcal{V})} f(S, \bar{w})p_t(S),$$ is maximized. We now consider the following strategy: in each round $t$, for any seed set $S \in \mathcal{P}(\mathcal{V})$, select it with probability $p(S) = \sum_{t = 1}^T p_t(S)/T$. The expected cost of this strategy is 
	\begin{equation*}
	\begin{split}
    T \cdot \sum_{S \in \mathcal{P}(\mathcal{V})}\mathbf{c}(S)p(S) 
    &= T \cdot \sum_{S \in \mathcal{P}(\mathcal{V})} \mathbf{c}(S) \cdot \sum_{t = 1}^T p_t(S)/T\\ 
    & = \sum_{t = 1}^T \sum_{S \in \mathcal{P}(\mathcal{V})} \mathbf{c}(S)p_t(S) \leq B.
	\end{split}
	\end{equation*}
	Thus this strategy respects the expected budget constraint. Furthermore, the expected reward of this strategy is 
	\begin{align*}
	&T \cdot \sum_{S \in \mathcal{P}(\mathcal{V})}f(S, \bar{w})p(S)\\ 
	= &T \cdot \sum_{S \in \mathcal{P}(\mathcal{V})} f(S, \bar{w}) \cdot \sum_{t = 1}^T p_t(S)/T\\ 
	= &\sum_{t = 1}^T \sum_{S \in \mathcal{P}(\mathcal{V})} f(S, \bar{w})p_t(S),
	\end{align*}
	which is equal to that of the optimal strategy. Finally, note that $p$ is a feasible solution to the LP1 with $b = B/T$.
	\qed
	
From Lemma \ref{lemma:optimbbar}, we can conclude that the optimal reward of Lin-IMB-L can be written as $\sum_{t=1}^T \E(f(S^*,\bar{w}))$, where $S^*$ is the seed set sampled from $\mathcal{P}(\mathcal{V})$ following the probability distribution $p^*$ defined in the lemma. 

For any online learning algorithm of Lin-IMB-L that employs an $(\alpha,\beta)$-approximation oracle to select seed set $S_t$ in each round $t$, the expected reward is $\sum_{t=1}^T \E(f(S_t,\bar{w}))$. The cumulative scaled regret, i.e., the optimal expected reward minus the expected reward of the online learning algorithm scaled up by $\eta = \alpha\beta$, is 
\begin{align*}
R^\eta(T) &= \sum_{t=1}^T \E(f(S^*,\bar{w})) - \sum\limits_{t=1}^T \E(f(S_t,\bar{w}))/\eta\\ 
&= \sum_{t=1}^T \E\Big[f(S^*,\bar{w}) - \frac{f(S_t,\bar{w})}{\eta}\Big].
\end{align*}

\subsection{Cumulative Scaled Regret Proxy} \label{apd:proxy}
Since solving for the optimal distribution $p^*$ for sampling $S^*$ is NP-hard, we cannot directly compute $R^\eta(T)$. 
Let $S^{ora}(w)$ be the seed set chosen by a randomized $(\alpha,\beta)$-approximation oracle with input edge weights $w$. In our numerical experiments, we compute the quantity
$\hat{R}^\eta(T) = \sum_{t = 1}^T \E(\hat{R}^\eta_t)$ instead, where
\begin{equation*}
\begin{split}
     \E(\hat{R}^\eta_t) &= \E \Big[f(S^{ora}(\bar w),\bar w)  -  f(S_t,\bar w) \Big], \;\; \eta = \alpha \beta.
\end{split}
 \end{equation*}

From the definition of $(\alpha,\beta)$-approximation oracle, we have that $$\mathbb{E}(f(S^{ora}(\bar w),\bar w)) \geq \eta \cdot \mathbb{E}(f(S^*,\bar w)).$$ As a result, $$\hat{R}^{\eta}(T)/\eta \geq R^{\eta}(T).$$ Therefore, the growth of the actual cumulative scaled regret $R^\eta(T)$ is upper bounded by a constant factor times that of $\hat{R}^\eta(T)$. We call $\hat{R}^\eta(T)$ the \emph{cumulative scaled regret proxy}. For our numerical experiments, we report $\hat{R}^\eta(T)$ instead of the true cumulative scaled regret in Figure \ref{fig:25-node} and \ref{fig:50-node}.

\section{Proofs of offline results}
\subsection{NP-Hardness} \label{apd:nphardness-proof} 
\begin{theorem}\label{thm:nphardness}IMB is NP-hard.\end{theorem} 
\textit{Proof of Theorem \ref{thm:nphardness}.}
		Given any instance of the \textit{minimum set cover problem} in the following form: 
		
		$U = \{u_1, u_2, ..., u_n\}$ is a ground set with $n$ elements. $\mathcal{S} = \{S_1, S_2, ..., S_m\}$ is a family of $m$ subsects of $U$. Find a minimal cardinality subset $\mathcal{S}'$ of $\mathcal{S}$ such that $\cup_{S_i \in \mathcal{S}'} S_i = U$. 
		
		Assume IMB can be solved efficiently. We show that the given minimum set cover instance can be solved efficiently. 
		
		First, construct a network $\cD = (\mathcal{V}, \mathcal{E})$ as follows. For each $S_i \in \mathcal{S}$, there is a node in $\mathcal{V}$ that corresponds to it. For each $u_j \in U$, there is a node that corresponds to it. $(S_i, u_j) \in \mathcal{E}$ if and only if $u_j \in S_i$.
		
		Use IMB$(\cD, b, \mathbf{c}, w)$-OPT to denote the optimal solution for an IMB instance with budget $b$, node costs $\mathbf{c}$ and edge weights $w$. Note that this solution can be expressed as a probability distribution on seed sets that specifies the likelihood with which each seed set will be played. 
		
		Find the smallest integer $k$ such that the expected number of activated nodes of IMB$(\cD, k, \mathbf{1}, \mathbf{1})$-OPT is at least $n+ k$. Note such a $k$ must exists and is smaller than $m$ since by our assumption, $\mathcal{S}$ covers $U$. Since IMB$(\cD, b, \mathbf{1}, \mathbf{1})$-OPT can be obtained efficiently for each $b \in \{0, 1, ..., m\}$, $k$ can be found efficiently.
		
		We claim that i) $k$ is the smallest size of $\mathcal{S}'$, that is, the smallest number of subsets needed to cover $U$; ii) any $S \subseteq \mathcal{V}$ with positive probability in IMB$(\cD, k, \mathbf{1}, \mathbf{1})$-OPT must correspond to a set cover for $U$ of size $k$. To prove ii), note that with cardinality constraint $k$, the maximum number of activated nodes in $\cD$ is $n+k$ due to the way we construct the network. Since the expected number of activated nodes of IMB$(\cD, k, \mathbf{1}, \mathbf{1})$-OPT is at least $n+k$, only $S \in \mathcal{V}$ such that $f(S, w) = n+ k$ can have positive probability. Without loss of generality, $|S| \leq k$. This is because if $|S| > k$, by cardinality constraint, there must exists a subset $S'$ with positive probability such that $|S'| < k$. Furthermore, $f(S, w) \leq n+ |S|$, and thus $|S| \geq k$. As a result, $|S| = k$. From ii), we know that we need at most $k$ subsets in $\mathcal{S}$ to cover $U$. If there exits a family $\mathcal{S}^*$ of $h$ subsets in $\mathcal{S}$ that covers $U$, where $h < k$, then $p(\mathcal{S}^*) = 1$ is a feasible solution to IMB$(\cD, h, \mathbf{1}, \mathbf{1})$ with objective value $n + h$. Thus, the expected number of activated nodes of IMB$(\cD, h, \mathbf{1}, \mathbf{1})$-OPT is at least $n + h$, contradicting the assumption that $k$ is the smallest integer such that the expected number of activated nodes of IMB$(\cD, k, \mathbf{1}, \mathbf{1})$-OPT is at least $n+ k$. i) therefore follows.\qed
		
	\subsection{Proof of Theorem \ref{offline_thm2}} \label{apd:offline_thm2-proof}
	We first study an alternative approximation oracle {\tt{ORACLE-IMB-a}} detailed below. We show that the distribution obtained in {\tt{ORACLE-IMB}} is an optimal solution to the LP in {\tt{ORACLE-IMB-a}} using Lemma \ref{lemma:2set}.
	\begin{algorithm}
\SetAlgoLined
  \KwData{$\cD = (\cV, \mathcal{E}), b, \mathbf{c}, w$}
  \KwResult{$S\subseteq \mathcal{V}$}
 initialization: $S_0 = \emptyset$\;
 \For{$i = 1, 2, ..., n$}{Compute $v_i = {\arg\max}_{v \in \mathcal{V}\backslash S_{i-1}} \frac{f(S_{i-1}\cup\{v\}, w)-f(S_{i-1}, w)}{\mathbf{c}(v)}$\;
 Set $S_i = S_{i-1} \cup \{v_i\}$\;}
 Solve the following LP to get an optimal solution $p^* = (p_0^*, p_1^*, ..., p_n^*)$:\\
$\max \; \sum\limits_{j = 0}^n p_j f(S_j, w)$\\ $\text{s.t}\; \sum_{j = 0}^n \mathbf{c}(S_j) p_j \leq b$;\\ 
$\;\;\;\;\;\sum_{j = 0}^n p_j = 1;$\\ 
$\;\;\;\;\;p_j \geq 0 \;\forall \; j = 0, 1, ..., n.$ \;
Sample $S$ from $\{S_0, S_1, ..., S_n\}$ with probability distribution $p^*$
 \caption{{\tt{ORACLE-IMB-a}}}
\end{algorithm}
	
	\begin{lemma} \label{lemma:2set}
There exits an optimal solution $p^*$ to the LP in {\tt{ORACLE-IMB-a}} that has the following properties: 1) at most two elements in $p^*=(p_0^*, p_1^*, ..., p_n^*)$ are non-zero; 2) if $p_i^*, p_j^*$ are non-zero, then $|i-j| \leq 1$.
\end{lemma}

\textit{Proof of Lemma \ref{lemma:2set}.}
For conciseness, we suppress $w$ as a argument of $f(S, w)$. Also, without loss of generality, assume $i < j$. Now suppose that there exists $p_i^*, p_i^*$ such that $|i-j| \geq 2$. Then the contribution of $S_i, S_j$ to the objective value of the LP is $p_i^*f(S_i) + p_j^*f(S_j)$ and the consumption of the budget is $p_i^*\mathbf{c}(S_i) + p_j^*\mathbf{c}(S_j)$. Now let $p_i', p_j'$ be the solution to the following system of equations:
        
\begin{equation}
\begin{split}
p_i' + p_j' &= p_i^* + p_j^*\\
p_i'\mathbf{c}(S_{i+1}) + p_j'\mathbf{c}(S_{j-1}) &= p_i^*\mathbf{c}(S_i) + p_j^*\mathbf{c}(S_j)
\end{split}
\end{equation}
        
The system of equations has a unique solution since $\mathbf{c}(S_{i}) \leq \mathbf{c}(S_{i+1}) \leq \mathbf{c}(S_{j-1}) \leq \mathbf{c}(S_{j})$. Note that $p_i'f(S_{i+1}) + p_j' f(S_{j-1}) \geq p_i^*f(S_i) + p_j^*f(S_j)$ by the observation that $f(S_k, w) - f(S_{k-1}, w) \geq f(S_l, w) - f(S_{l-1}, w)$ for $1 \leq k \leq l \leq n$. Therefore, allocating $p_i'$ to $S_{i+1}$ and $p_j'$ to $S_{j-1}$ costs the same as $p_i^*$ to $S_{i}$ and $p_j^*$ to $S_{j}$, but it achieves at least the same expected influence spread.
        
By repeating the process, eventually at most two consecutive sets in $\{S_0, S_1, ..., S_n\}$ will have positive probability.\qed

Since the expected cost is at most $b$, by Lemma \ref{lemma:2set}, we know that if two sets in $\{S_0, S_1, ..., S_n\}$ have positive probability, then one of them is the largest set whose cost is below $b$, denoted by $S_{-}$, and the other one is the smallest set whose cost is greater than $b$, denoted by $S_+$. As a result, in the for loop of {\tt{ORACLE-IMB-a}}, we can stop as soon as $\mathbf{c}(S_i) \geq b$. When solving the LP, instead of having $n+1$ variables, we solve for the optimal probability distribution on $S_-$ and $S_+$ only. This simplified algorithm is what we presented as {\tt{ORACLE-IMB}}.
	
	\textit{Proof of Theorem \ref{offline_thm2}.}
	
		To prove Theorem \ref{offline_thm2}, we first observe that IMB can be equivalently formulated as LP1 in \eqref{lp1}.
		
		We also need some definitions and lemmas.
	\begin{definition}
		For any budget $b$ and edge weights $w$, let $S(b,w) := {\arg\max}_{S: \mathbf{c}(S) \leq b} f(S, w)$, and call it the \textit{best response set} of budget $b$ and influence $w$.  
	\end{definition}
	\begin{definition}
	
		A family of seed sets $$\mathcal{S'}=\{S'_1,S'_2,S'_3,\cdots,S'_L\}$$  is called a \textit{$\beta$-approximation family} \textit{with respect to $w$} if for any $b$ and its corresponding best respond set $S(b, w)$, we can find a seed set $S'_i$ in $\mathcal{S'}$ such that $\mathbf{c}(S'_i) \le b$ and $f(S'_i, w) \ge \beta f(S(b,w), w)$.
	\end{definition}
	
	\begin{definition}
		A family of seed sets $$\mathcal{S'}=\{S'_1,S'_2,S'_3,\cdots,S'_L\}$$  is called a \textit{$\beta$-combo-approximation family} \textit{with respect to $w$} if for any budget $b$ and its corresponding best respond set $S(b, w)$, we can find two seed sets $S'_{i},S'_{j}$ in $\mathcal{S'}$ and probabilities $q_{i}+q_{j}=1$ such that $q_{i}\mathbf{c}(S'_{i})+ q_{j}\mathbf{c}(S'_{j}) \le b$ and $q_{i}f(S'_{i}, w) + q_{j}f(S'_{j}, w) \ge \beta f(S(b,w), w)$.
	\end{definition}
	
	\begin{lemma} \label{lemma:combo-family}
		If a family of seed sets $$\mathcal{S'}=\{S'_1,S'_2,S'_3,\cdots,S'_L\}$$ is a $\beta$-approximation family \textit{or} a $\beta$-comb-approximation family with respect to to $w$, then for any budget $b$, there exists a probability distribution $p'=(p'_1,p'_2,p'_3,\cdots,p'_L)$ over $\mathcal{S'}$ such that 
		$\sum_{i = 1}^L p'_i \mathbf{c}(S'_i) \leq b$ and 
		\begin{align*}\sum_{i = 1}^L  f(S'_i, w) p'_i &\geq \beta \sum_{S \in P(\mathcal{V})} f(S, w) p^{opt}(S)\\ &= \beta \cdot \E(f(S^*,  w)),\end{align*}
		where $p^{opt}$ is a probability distribution over all $|P(\mathcal{V})|$ possible seed sets used by an optimal oracle for IMB (i.e., $p^{opt}$ is an optimal solution to LP1), and $S^*$ is the seed set returned by an optimal oracle.
	\end{lemma}
	\textit{Proof of Lemma \ref{lemma:combo-family}.}
	Initialize $p'_i = 0 \;\; \forall i = 1, ..., L$. For each $S \in P(\mathcal{V})$, from the definition of $\beta$-(comb)-approximation family with respect to to $w$, there exist two seed sets $S'_i, S'_j$ in $\mathcal{S'}$ together with probabilities $q_i, q_j$ such that $q_{i}+q_{j}=1$, $q_{i}\mathbf{c}(S'_{i})+ q_{j}\mathbf{c}(S'_{j}) \le \mathbf{c}(S)$ and $q_{i}f(S'_{i}, w) + q_{j}f(S'_{j}, w) \ge \beta f(S(b,w), w) \geq \beta f(S, w)$. Update $p'_i \mapsto p'_i+p^{opt}(S)q_i$, $p'_j \mapsto p'_j+p^{opt}(S)q_j$. After doing so for all $S \in P(\mathcal{V})$, we have the probability distribution $p'$ as desired. \qed

	To prove Theorem \ref{offline_thm2}, we show that $\mathcal{S} = \{S_0, S_1, ..., S_n\}$ as constructed in {\tt{ORACLE-IMB-a}} is a ($1-1/e$)-comb-approximation family with respect to $w$.
    As {\tt{ORACLE-IMB-a}} solves an LP to find the optimal distribution $p^*$ over $\mathcal{S}$, it indeed achieves an $1-1/e$ approximation ratio.
	
	Consider the sequence of sets, $S_0, S_1, ..., S_n$, constructed in the oracle. For any budget $0 < b \leq \mathbf{c}(\mathcal{V})$, we can find a unique index $i(b) \in \{1, 2, ..., n\}$ such that $\mathbf{c}(S_{i(b)-1}) < b \leq \mathbf{c}(S_{i(b)})$, and a unique $\alpha \in (0,1]$ such that $b = (1-\alpha)\mathbf{c}(S_{i(b)-1}) + \alpha \mathbf{c}(S_{i(b)})$. 
	We now only need to show that 
	\begin{align*}
	&(1-\alpha)f(S_{i(b)-1}, w) + \alpha f(S_{i(b)}, w)\\ \geq &\Big( 1- e^{-1}\Big)f(S(b,w), w).
	\end{align*}
	
	Define $$r_i = \max_{v \in \mathcal{V} \backslash  S_{i-1}} \frac{f(S_{i-1}\cup\{v\}, w)-f(S_{i-1}, w)}{\mathbf{c}(v)}.$$ Let $x_0=0, x_j=\mathbf{c}(S_{j})$ for $j=1,2,\cdots,i(b)-1$, and $x_{i(b)}=b$.
	We define a density function $p(x)$ on $[0,B]$ as $p(x):=r_{j+1}$ if $x \in [x_{j}, x_{j+1})$.
	We denote $h(x):=\int_0^{x} p(s) ds$.
	
	Now as $f(\cdot, w)$ is submodular and by the definition of $r_{j+1}$, we have that
	$f(S_j,\omega)=h(x_j)$ for $j=1,2,\cdots,i(b)-1$, and
    \begin{equation}\label{eq:ORACLE-IMcB-1}
    \begin{split}
	f(S(b,w), w) \le &f(S(b,w) \cup S_j, w)\\ \le& f(S_j, w)+ b \cdot r_{j+1}
	\end{split}
	\end{equation}
	for $0 \le j \le i(b)-1.$
	
	\eqref{eq:ORACLE-IMcB-1} can be relaxed to
	\begin{equation*}
		f(S(b,w), w) \le h(x)+b \cdot h'(x)\;\; \forall x \in [0,b].
	\end{equation*}
	
	Thus, we have 
	\begin{align*}
		e^{\frac{x}{b}} f(S^*) 
		&\le  e^{\frac{x}{b}} (h(x)+b \cdot h'(x))\\
		&= (e^{\frac{x}{b}} h(x))' b
	\end{align*}
	for $x \in [0,b]$.
	With the initial conditions $h(0)=0$ and $h'(0)=p(0)>0$, we get that 
	$$e^{\frac{x}{b}} h(x) \ge \int_0^x \frac{e^{\frac{s}{b}}}{b} f(S^*) ds 
	=(e^{\frac{x}{b}}-1) f(S^*).$$
	Taking $x=b$, we have that 
	\begin{align*}
	h(b) \ge (1-e^{-\frac{b}{b}})f(S^*) >(1-e^{-1})f(S^*).  
	\end{align*}
	Recall that 
	$b = (1-\alpha)\mathbf{c}(S_{i(b)-1}) + \alpha \mathbf{c}(S_{i(b)})$. Therefore, we have 
	\begin{align*}
	h(b)&=\int_0^{b} p(s) ds\\
	&=\int_0^{\mathbf{c}(S_{i(b)-1})} p(s) ds+\int_{\mathbf{c}(S_{i(b)-1})}^{b} p(s) ds\\
	&=\int_0^{\mathbf{c}(S_{i(b)-1})} p(s) ds+(b-\mathbf{c}(S_{i(b)-1}))r_{i(b)}\\
	&=(1-\alpha) f(S_{i(b)-1}, w)+\alpha f(S_{i(b)}, w)
	\end{align*} as desired.  \qed
	
\section{Simulation of $f(\cdot, w)$ and Modified Approximation Oracle} \label{apd:simulation}

	In the main body of the paper, we give oracles for IMB with the assumption that $f(\cdot, w)$ can be computed exactly. However, it is \#P-hard to compute this quantity \citep{Chen:2010:SIM:1835804.1835934}, and thus we need to approximate it by simulation. In \cite{Kempe:2003}, the authors propose to simulate the random diffusion process and use the empirical mean of the number of activated users to approximate the expected influence spread. In their numerical experiments, they use 10,000 simulations to approximate $f(S, w)$ for each seed set $S$. Such a method greatly increases the computational burden of the greedy algorithm. \cite{DBLP:journals/corr/abs-1212-0884} propose a very different method that samples a number of so-called  \textit{Reverse Reachable (RR) sets} and use them to estimate influence spread under the IC model. %, and reduce the IM problem to the \textit{maximum set coverage problem} \citep{Vazirani:2001}. 
	
	Based on the theoretical breakthrough of \cite{DBLP:journals/corr/abs-1212-0884}, \cite{Tang:2015:IMN:2723372.2723734,Tang:2014:IMN:2588555.2593670} present \emph{Two-phase Influence Maximization (TIM)} and \emph{Influence Maximization via Martingales (IMM)} for IM with complexity $O((m+n)K \epsilon^{-2}\log(n)))$, where $m$ is the number of edges in the network, $n$ the number of nodes, $K$ the cardinality constraint of the seed sets, and $\epsilon \in (0,1)$ the size of the error. These two methods improve upon the algorithm in  \cite{DBLP:journals/corr/abs-1212-0884} that has a run time complexity of $O((m+n)K \epsilon^{-3}\log(n)))$. 
	%Both TIM and IMM have a parameter estimation phase and a node selection phase, the latter of which samples numerous node sets from the network to find a solution for influence maximization. 
	All these three simulation-based methods are designed solely for IM with simple cardinality constraints. Their analysis relies on the assumption that the optimal seed set is of size $K$. As a result, the number of RR sets required in their methods does not guarantee estimation accuracy of $f(S, w)$ for seed sets of bigger sizes. However, in our problems, the feasible seed sets can potentially be of any sizes. In particular, our {\tt{ORACLE-IMB}} assigns a probability distribution to seed sets of cardinalities from $0$ to $n$. This means that our simulation method needs to guarantee accuracy for seed sets of all sizes.

	In order to cater to this requirement of our budgeted problems, we extend the results in \cite{DBLP:journals/corr/abs-1212-0884} and \cite{Tang:2014:IMN:2588555.2593670}, and develop a \textit{Concave Error Interval (CEI) analysis} which gives an upper bound on the number of RR sets required to secure a consistent influence spread estimate for seed sets of different sizes with high probability. 
	We then detail how we modify our oracles using RR sets to estimate $f(\cdot, w)$. We prove $1-1/e-\epsilon$-approximation guarantees for the modified oracles. We also supply the run time complexity analysis. In the rest of the section, we suppress $w$ as an argument of $f(\cdot, w)$. %{\color{red}(Modify after going through the rest of the section)}
	
	\subsection{Reverse Reachable (RR) Set}

	To precisely explain our simulation method, we state the formal definition of RR sets introduced by \cite{Tang:2014:IMN:2588555.2593670}.
	\begin{definition}[Reverse Reachable Set]
		Let $v$ be a node in $\mathcal{V}$, and $\mathcal{H}$ be a graph obtained by removing each directed edge $e$ in $\mathcal{E}$ with probability $1-p(e)$. 
		The reverse reachable (RR) set for $v$ in $\mathcal{H}$ is the set of nodes in $\mathcal{H}$ that can reach $v$. 
		That is, a node $u$ is in the RR set if and only if there is a directed path from $u$ to $v$ in $\mathcal{H}$.
	\end{definition}
	
	\begin{definition}[Random RR Set]
		\label{def:RR-set}
		Let $\mathcal{W}$ be the distribution on $\mathcal{H}$ induced by the randomness in edge removals from $\mathcal{V}$. 
		A random RR set is an RR set generated on an instance of $\mathcal{H}$ randomly sampled from $\mathcal{W}$, for a node selected uniformly at random from $\mathcal{V}$.
	\end{definition}
	
\cite{Tang:2014:IMN:2588555.2593670} give an algorithm for generating a random RR set, which is presented in Algorithm \ref{alg:RRset} below. 

\begin{algorithm} 
\SetAlgoLined
  \KwData{digraph $\cD=\mathcal{(V, A)}$, edge weights $w: \mathcal{E} \mapsto [0,1]$}
  \KwResult{Random RR set $R$}
 initialization: $R=\emptyset$, first-in-first-out queue $Q$\;
 sample a node $v$ uniformly at random from $\mathcal{V}$, add to $R$\;
 \For{$u \in \mathcal{V}$ s.t. $(u, v) \in \mathcal{E}$}{
 flip a biased coin with probability $w(u,v)$ of turning head\;
 \If{the coin turns head}{Add $u$ to $Q$ and $R$}
 }
 \While{$Q$ is not empty}{
 extract the node $v'$ at the top of $Q$\;
 \For{$u' \in \mathcal{V}$ s.t. $(u', v') \in \mathcal{E}$}{
 flip a biased coin with probability $w(u',v')$ of turning head\;
 \If{the coin turns head}
 {add $u'$ to $Q$ and $R$}
 }
 }
\caption{Random RR set} \label{alg:RRset}
\end{algorithm}
	
	By \cite{DBLP:journals/corr/abs-1212-0884}, we have the following lemma:
	\begin{lemma}[\citeauthor{DBLP:journals/corr/abs-1212-0884}]
		\label{lemma:cover} 
		For any seed set $S$ and node $v$, the probability that a diffusion process from $S$ which follows the IC model can activate $v$ equals the probability that $S$ overlaps an RR set for $v$ in a graph $\mathcal{H}$ generated by removing each directed edge $e$ in $\mathcal{E}$ with probability $1-p(e)$.
	\end{lemma}
	Suppose we have generated a collection $\RR$ of random RR sets.
	For any node set $S$, let $F_{\RR}(S)$ be the fraction of RR sets in $\RR$ that overlap $S$.
	From Lemma~\ref{lemma:cover}, \cite{Tang:2014:IMN:2588555.2593670} showed that the expected value of $n F_{\RR}(S)$ equals the expected influence spread of $S$ in $\mathcal{V}$, i.e., $\E[n F_{\RR}(S)] =f(S)$.
	Thus, if the number of RR sets in ${\RR}$ is large enough, then we can use the realized value $nF_{\RR}(S)$ to approximate $f(S)$. 
	
	\subsection{Concave Error Interval and Simulation Sample Size}
	% To secure the accuracy, if in each round, the simulation collection ${\RR}$ can give us estimates of \emph{every} $F_{\RR}(S)$ within small error with high probability, then we  can estimate $f(S)$ relatively accurate for any seed set $S$ in both IMrB and IMcB.
	In this section, we propose the Concave Error Interval (CEI) method of analysis which gives the number of random RR sets required to obtain a close estimate of $f(S)$ using $nF_{\RR}(S)$ for every seed set $S$.
	Our analysis uses the following Chernoff inequality.
	\begin{lemma}[Chernoff Bound]
		\label{lemma:Chernoff}
		Let $X$ be the sum of $L$ i.i.d random variables sampled from a distribution on $[0,1]$ with a mean $\mu$. 
		For any $\eta>0$,
		\begin{align*}
			& \pr(X/L-\mu \ge \eta \mu) \le e^{-\frac{\eta^2}{2+\eta} L\mu},\\
			& \pr(X/L-\mu \le -\eta \mu) \le e^{-\frac{\eta^2}{2+\eta} L\mu}.
		\end{align*}
	\end{lemma}
	
	\subsubsection{Concave Error Interval}
	\textbf{}
	
	Let $OPT^{B}=\sum\limits_{S} p^{opt}(S)f(S)$ be the expected influence spread of {\tt{OPTIMAL-IMB}}, the optimal stochastic strategy for IMB; let $OPT^{K}$ be the expected influence spread of the optimal seed set for IM with cardinality constraint $K$. 
	%The \emph{TIM} algorithm in \cite{Tang:2014:IMN:2588555.2593670} is for IM with the cardinality constraint, and their analysis relies on the fact that the expected influence spread of the seed set output by TIM is always smaller than $OPT^{K}$. However, our {\tt{OPTIMAL-IMB}} and {\tt{ORACLE-IMB}} both place a probability distribution on seed sets of all possible sizes, including ones whose expected influence spread is larger than that of the optimal.  As a result, the analysis in \cite{Tang:2014:IMN:2588555.2593670} won't go through in our budgeted problems.
	
	We now introduce a \textit{concave error interval $I_{S}$} for each seed set $S$, and define an event $E_{S}$ as follows which limits the difference between $n F_{\RR}(S)$ and $f(S)$. 
	Suppose $\epsilon$ is given. 
	\begin{definition}[Concave Error Interval $I_{S}$ and Event $E_{S}$]
		\begin{align*}
			I_{S}= \Big[&-\frac{\epsilon}{1+\sqrt{1-1/e}} \sqrt{OPT^{B} f(S)},\\
			&\frac{\epsilon}{1+\sqrt{1-1/e}} \sqrt{OPT^{B} f(S)}\Big],\\
			E_{S}=\Big\{&n F_{\RR}(S)-f(S) \in  I_{S}\Big\}.
		\end{align*}
	\end{definition}
	The length of the error interval $I_{S}$ is $$\frac{2\epsilon}{1+\sqrt{1-1/e}}  \sqrt{OPT^{B} f(S)},$$ which is concave in $f(S)$.
	$E_{S}$ is the event that the difference between $n F_{\RR}(S)$ and its mean $f(S)$ is within $I_{S}$.
	With $L$ being the number of randomly sampled RR sets, the likelihood of $E_{S}$ can be bounded as follows.
	\begin{align*}
		 \pr(E_{S}) 
		= \pr \Big(&|n F_{\RR}(S)-f(S)|\\ &\le \frac{\epsilon}{1+\sqrt{1-1/e}}  \sqrt{OPT^{B} f(S)}\Big) \\ \nonumber
		= \pr \Big(&|L F_{\RR}(S)- L \frac{f(S)}{n} | \\ &\le  \sqrt{\frac{OPT^{B}}{f(S)}} \frac{\epsilon}{1+\sqrt{1-1/e}}  L \frac{f(S)}{n}\Big).
	\end{align*}
	Let $$\eta=\sqrt{\frac{OPT^{B}}{f(S)}} \frac{\epsilon}{1+\sqrt{1-1/e}}.$$ 
	By Lemma~\ref{lemma:Chernoff}, we have that when $\epsilon \le {3}/{\sqrt{n}}$, 
	\begin{align*}
	\pr(E_{S}) \ge & 1-2 e^{-\frac{\eta^2}{2+\eta} L \frac{f(S)}{n}}\\ \ge & 1-2 e^{\frac{OPT^{B}L}{3n}(\frac{\epsilon}{1+\sqrt{1-1/e}})^2}.
	\end{align*}
	
	Therefore, we have a uniform lower bound on $\pr(E_{S})$ for every seed set $S$, which implies the following lemma:
	\begin{lemma}
		\label{lemma:one-set}
		For any given $l$, let
		\begin{align} \label{eq:L}
			L=\frac{7n(l \log n+ n \log 2)}{OPT^{B} \cdot \epsilon^2}
		\end{align}
		If $\RR$ contains $L$ random RR sets and $\epsilon \le {3}/{\sqrt{n}}$, then for every seed set $S$ in $\mathcal{V}$, $E_{S}$ happens with probability at least $1-{1}/{n^{l} 2^{n}}.$
	\end{lemma}
	
	Since there are $2^n$ different seed sets, we have the following.
	\begin{lemma}
		\label{lemma:all}
		For any given $l$ and its corresponding $L$ defined in \eqref{eq:L},
		 if $\RR$ contains $L$ random RR sets and $\epsilon \le {3}/{\sqrt{n}}$, then
		\begin{align*}
			\pr \big( E_{S} \textrm{ holds for all } S \big) > 1-\frac{1}{n^{l}}.
		\end{align*}
	\end{lemma}
	
	So far, we have established the relationship between the number of random RR sets and the estimation accuracy through the CEI analysis. %Thus, we can use $n F_{\RR}(S)$ to approach $f(S)$, and we call it the RR sets influence simulation technique.
	Later we will prove that  {\tt{ORACLE-IMB}} combined with the above RR sets simulation technique give at least $(1-1/e-\epsilon)$-approximation guarantee for IMB with high probability.
	
	In \cite{Tang:2014:IMN:2588555.2593670}, it is shown that \emph{TIM} returns an $(1-1/e-\epsilon)$-approximation solution with an expected runtime of $O({(k+l)(m+n) \log n}/{\epsilon^2})$, which is near-optimal under the IC diffusion model, as it is only a $\log n$ factor away from the lower-bound established by \cite{DBLP:journals/corr/abs-1212-0884}.
	As will be shown later, the expected runtime of the modified {\tt{ORACLE-IMB}} is $O({m(l \log n+ n)}/{\epsilon^2})$, which has an extra $n$ compared to the lower-bound due to the flexible usage of the total budget. 
	However, we give guaranteed influence spread estimates for all $2^n$ possible seed sets, while the TIM analysis only covers the $\binom{n}{K}$ size-K seed sets. 
	Under the $\log$ operator, the difference in runtime is $n \log 2$ versus $k \log n$.
	
	\subsection{$(1-1/e-\epsilon)$-Approximation Ratio for  {\tt{ORACLE-IMB-M}}}
	
	We denote by {\tt{ORACLE-IMB-M}} the modified version of {\tt{ORACLE-IMB}} that includes the $f(S)$ approximation. Assume $l$ and $\epsilon \leq 3/\sqrt{n}$ are given.
	
	\begin{algorithm} 
\SetAlgoLined
  Generate a collection $\RR$ of $L$ random RR sets where $L$ is as defined in \eqref{eq:L}\;
  Run {\tt{ORACLE-IMB}} with the change that whenever a f(S) needs to be computed, use $nF_{\RR}(S)$ instead
 \caption{{\tt{ORACLE-IMB}-M}}
 \label{cm-m}
\end{algorithm}
	
	To prove the approximation guarantee for {\tt{ORACLE-IMB-M}}, we need the following theorem. 
	
	\begin{theorem}
		\label{thm:k}
		Let $l$ and $\epsilon \leq 3/\sqrt{n}$ be given and $L$ be as defined in \eqref{eq:L} with respect to $l$. For any stochastic strategy in the form of a probability distribution $p(S)$ over a family of seed sets $S$, assume that $n F_{\RR}(S)$ is used to approximate $f(S)$ where $\RR$ is a collection of $L$ randomly sampled RR set from Definition~\ref{def:RR-set}.
		We have that with probability at least $1-1/n^l$,
		\begin{equation}
			\label{eq:Fgef}
			\begin{split}
			&\sum\limits_{S} p(S) n F_{\RR}(S)\\ \ge & \sum\limits_{S} p(S)f(S) \\ & - \frac{\epsilon}{1+\sqrt{1-1/e}}  \sqrt{OPT^{B} \sum\limits_{S} p(S)f(S)},
			\end{split}
		\end{equation}
		and 
		\begin{equation}
			\label{eq:fgeF}
			\begin{split}
			&\sum\limits_{S} p(S) f(S)\\  \ge & \sum\limits_{S} p(S)n F_{\RR}(S) \\ &- \frac{\epsilon}{1+\sqrt{1-1/e}}  \sqrt{OPT^{B} \sum\limits_{S} p(S)n F_{\RR}(S)}.
			\end{split}
		\end{equation}
	\end{theorem}
	\textit{Proof of Theorem~\ref{thm:k}:}
	By Lemma~\ref{lemma:all}, we have that with probability $1-\frac{1}{n^{l}}$, for all $S$ 
	\begin{align*}
		n F_{\RR}(S) \ge f(S)-\frac{\epsilon}{1+\sqrt{1-1/e}}  \sqrt{OPT^{B} f(S)}.
	\end{align*}  
	Since $\sqrt{OPT^{B} f(S)}$ is concave in $f(S)$, using Jensen's inequality we have 
	\begin{align*}
		& \sum\limits_{S} p(S)n F_{\RR}(S) \nonumber\\
		\ge &\sum\limits_{S} p(S)\Big(f(S)-\frac{\epsilon}{1+\sqrt{1-1/e}}  \sqrt{OPT^{B} f(S)}\Big)\\\nonumber
		\ge &\sum\limits_{S} p(S)f(S) \\&- \frac{\epsilon}{1+\sqrt{1-1/e}}  \sqrt{OPT^{B} \sum\limits_{S} p(S)f(S)}.
	\end{align*} 
	
	Similarly, we have \begin{align*}
		n F_{\RR}(S) \le f(S)+\frac{\epsilon}{1+\sqrt{1-1/e}}  \sqrt{OPT^{B} f(S)}.
	\end{align*}
	By Jensen's inequality,
	\begin{align*}
		& \sum\limits_{S} p(S)n F_{\RR}(S) \nonumber\\
		&\le \sum\limits_{S} p(S)\Big(f(S) +\frac{\epsilon}{1+\sqrt{1-1/e}}  \sqrt{OPT^{B} f(S)}\Big)\\
		&\le \sum\limits_{S} p(S)f(S)+ \frac{\epsilon}{1+\sqrt{1-1/e}} \sqrt{OPT^{B} \sum\limits_{S} p(S)f(S)}.
	\end{align*}
	\qed
	
	We now prove the $1-1/e-\epsilon$-approximation guarantee for {\tt{ORACLE-IMB-M}}.
	\begin{theorem}
		\label{thm:k2}
		With probability at least $1-\frac{1}{n^{l}}$, the expected influence spread of the seed set returned by {\tt{ORACLE-IMB-M}} is at least $(1-1/e-\epsilon)$ that of the optimal spread.
	\end{theorem}
	\textit{Proof of Theorem~\ref{thm:k2}:}
	Let $p^{opt}(S)$ be the probability of selecting seed set $S$ for any $S \subseteq \mathcal{V}$ in {\tt{OPTIMAL-IMB}} when assuming $f(S)$ can be computed exactly. 
	Let $p^*$ be the probability distribution over seed sets computed in {\tt{ORACLE-IMB}-M} where $f(S)$ is approximated by RR sets.
    Since $F_{\mathcal{R}}(\cdot)$ is submodular, Theorem~\ref{offline_thm2} implies that
	\begin{align}
		\label{eq:1-1/e F}
		\sum\limits_{S} p^*(S) n F_{\RR}(S) \ge (1-1/e) \sum\limits_{S} p^{opt}(S) n F_{\RR}(S).
	\end{align}
	Now plugging $p^{opt}(S)$ into \eqref{eq:Fgef}, we get
	\begin{align}
		\label{eq:Ftof}
		&\sum\limits_{S} p^{opt}(S) n F_{\RR}(S)\\ \nonumber
		\ge & \sum\limits_{S} p^{opt}(S)f(S) \\ &\;\;- \frac{\epsilon}{1+\sqrt{1-1/e}}  \sqrt{OPT^{B} \sum\limits_{S} p^{opt}(S)f(S)}\\\nonumber
		= &OPT^{B}-\frac{\epsilon}{1+\sqrt{1-1/e}} OPT^{B}\\
		=&(1-\frac{\epsilon}{1+\sqrt{1-1/e}})OPT^{B}.
	\end{align}
	Furthermore, by plugging $p^*(S)$ into \eqref{eq:fgeF}, we get
	\begin{equation}
		\label{eq:ftoF}
		\begin{split}
		&\sum\limits_{S} p^*(S) f(S)\\ \ge & \sum\limits_{S} p^*(S)n F_{\RR}(S)\\  & \;\;-\frac{\epsilon}{1+\sqrt{1-1/e}}  \sqrt{OPT^{B} \sum\limits_{S} p^*(S)n F_{\RR}(S)}.
		\end{split}
	\end{equation}
	\eqref{eq:1-1/e F} \eqref{eq:Ftof} and \eqref{eq:ftoF} together give us that 
	\begin{align*}
		& \sum\limits_{S} p^*(S)f(S) 
	\\	\ge &\Big(1-\frac{\epsilon}{1+\sqrt{1-1/e}} \sqrt{(1-1/e)(1-\frac{\epsilon}{1+\sqrt{1-1/e}} )}\Big)\\ 
	& \cdot (1-1/e)(1-\frac{\epsilon}{1+\sqrt{1-1/e}})OPT^{B} \\
		\ge &(1-1/e-\epsilon)OPT^{B},
	\end{align*}
	with probability at least $1-{1}/{n^{l}}$, which completes the proof. \qed
	
	\subsection{Runtime Complexity of {\tt{ORACLE-IMB}-M}}
	The running time of {\tt{ORACLE-IMB}-M} mostly falls on generating random RR sets. To analyze the corresponding time complexity, we first define \textit{expected coin tosses (EPT)}.
	\begin{definition}
		$EPT$ is the expected number of coin tosses required to generate a random RR set following Algorithm \ref{alg:RRset}. 
	\end{definition}
	With the definition above, the expected runtime complexity of {\tt{ORACLE-IMB}-M} is $O(L\cdot EPT)$, where $L$ is the number of random RR sets required by the algorithm.  
	\cite{Tang:2014:IMN:2588555.2593670} establishes a lower bound of $OPT^{k}$ based on $EPT$. We bound $OPT^{B}$ similarly in the following lemma.
	\begin{lemma}
		\label{lemma:OPT-EPT}
		\begin{align*}
			\min(\frac{b}{\bar{c}},1)OPT^{B} \ge \frac{n}{m} EPT,
		\end{align*} 
		where $b$ is the budget and $\bar{c}$ is the maximum cost among all nodes.
	\end{lemma}
	\textit{Proof.}
	Let $R'$ be a random RR set, and let $p_{R'}$ be the probability that a randomly selected edge from $\cD$ points to a node in $R'$. 
	Then, $EPT=\E[p_{R'} \cdot m]$, where the expectation is taken over the random choices of $R'$.
	Let $Y(v,R')$ be a boolean function that returns 1 if $v \in R'$, and 0 otherwise.
	Denote $deg(v)$ as the in-degree of node $v$ in $\cD$ and $deg=\sum\limits_{v}deg(v)$.
	Then 
	\begin{align*}
		\frac{EPT}{m}=\E[p_{R'}]&=\sum\limits_{R' \in \RR} \pr(R') \cdot p_{R'}\\
		&=\sum\limits_{R' \in \RR} \pr(R') \cdot (\sum\limits_{v \in \mathcal{V}} \frac{deg(v)}{deg} Y(v,R'))\\
		&=\sum\limits_{v \in \mathcal{V}} \frac{deg(v)}{deg} \cdot (\sum\limits_{R' \in \RR}   \pr(R')  Y(v,R'))\\
		&=\sum\limits_{v \in \mathcal{V}} \frac{deg(v)}{deg} \cdot p_{v},
	\end{align*}
	where, by Lemma~\ref{lemma:cover}, $p_v=\sum\limits_{R' \in \RR}  \pr(R')  Y(v,R')$ equals the probability that a randomly selected node is activated given $v$ is in the seed set.
	Now consider a very simple policy $\Pi^{one}$ that selects one node $v$ as the seed set with probability ${deg(v)}/{deg}$. 
	Then ${n \cdot EPT}/{m}$ is the average expected influence of $\Pi^{one}$.
	It's easy to show that $$\min(\frac{b}{\bar{c}},1) OPT^{B} \ge f(\Pi^{one})=\frac{n\cdot EPT}{m},$$ where $f(\Pi^{one})$ is the expected influence spread of the seed set returned by policy $\Pi^{one}$.
%	\Halmos
	\qed
	
	Moreover, $EPT$ can be estimated by measuring the average width of RR sets, which is defined to be the average number of edges connecting to at least one node in a random RR set.
	If we choose
	$$L=\frac{7n(l \log n+ n\log 2)}{OPT^{B} \cdot \epsilon^2},$$ then the complexity of {\tt{ORACLE-IMB}} is $$O(L\cdot EPT)=O(\frac{m(l \log n+ n)}{\epsilon^2}),$$ since by Lemma \ref{lemma:OPT-EPT}, $$L \leq \frac{7m(l \log n+ n \log 2)\min(\frac{b}{\bar{c}},1)}{EPT \cdot \epsilon^2}.$$ Further note that the $OPT^{B}$ in the denominator of $L$ is not readily available. To closely approximate $L$, we apply the idea by \cite{Tang:2014:IMN:2588555.2593670}. We  compute $$L' = \frac{7n(l \log n+ n\log 2)}{n \cdot \epsilon^2},$$ which is a lower bound of $L$. We generate $L'$ many RR sets and then estimate $EPT$ using the average width of the generated RR sets, $\hat{EPT}$. If $$L' \leq \frac{7m(l \log n+ n \log 2)\min(\frac{b}{\bar{c}},1)}{\hat{EPT} \cdot \epsilon^2},$$ we keep generating RR sets to reach the quantity specified by the right-hand side of the proceeding inequality, and update our estimate of $EPT$ with all the available RR sets. We repeat this process until the current number of RR sets exceeds $$\frac{7m(l \log n+ n \log 2)\min(\frac{b}{\bar{c}},1)}{\hat{EPT} \cdot \epsilon^2}.$$
\end{document}